\pdfoutput=1   
\documentclass[letterpaper]{article} 
\usepackage[preprint]{varrate}  
\usepackage[hyphens]{url}  
\usepackage{graphicx} 
\usepackage{natbib}  
\usepackage{caption} 
\usepackage{algorithm}
\usepackage{algorithmic}

\usepackage{newfloat}
\usepackage{listings}
\DeclareCaptionStyle{ruled}{labelfont=normalfont,labelsep=colon,strut=off} 
\floatstyle{ruled}
\newfloat{listing}{tb}{lst}{}
\floatname{listing}{Listing}

\usepackage{booktabs}
\usepackage{adjustbox}
\usepackage{amsmath}
\usepackage{amssymb}
\title{VarRate: Training-Free Variable-Rate KV Cache Compression for Long-Context LLMs}
\author{
    Shahrzad Esmat,
    Dhawal Shah,
    Ali Jannesari
}
\affiliations{
    Iowa State University\\
    sesmat@iastate.edu, dhawal04@iastate.edu, jannesar@iastate.edu
}

\begin{document}

\maketitle

\begin{abstract}
The key--value (KV) cache is the main memory bottleneck in long-context large
language model (LLM) inference. Two leading training-free families are both
structurally limited: token-\emph{selection} methods (SnapKV, Ada-KV) score
importance from an observation window and \emph{evict} low-scoring tokens, but
eviction is irreversible---so when the importance signal degrades under
query-agnostic reuse, accuracy collapses by 11--15 points; uniform
\emph{low-rank} coding keeps every token but spends equal rank everywhere,
wasting budget. We observe that both failures share one cure: rank should be
\emph{allocated}, not evicted. We present \textbf{VarRate}, a training-free KV
codec that assigns each token a variable low-rank budget by its query salience,
keeping every token at a nonzero rank. Comparable adaptive-rank codecs reach
this allocation only through training; VarRate requires none. Because no token
is dropped, it degrades by only 3.5--5.5 points where query-aware selection collapses. At a
matched 20\% budget on LongBench (16 tasks), VarRate stays within 0.8 points of
the uncompressed model on both Llama-3.1-8B and Qwen2.5-7B. Averaged over the
two, it is the strongest matched-memory compressor. It significantly beats its
uniform-rank ablation on both models. Against KVzip, a method purpose-built for
query-agnostic reuse, it is accuracy-equivalent in three of four settings and
within a point overall, at about one-eighth the prefill overhead.
\end{abstract}


\begin{figure*}[t]
\centering
\includegraphics[width=\textwidth]{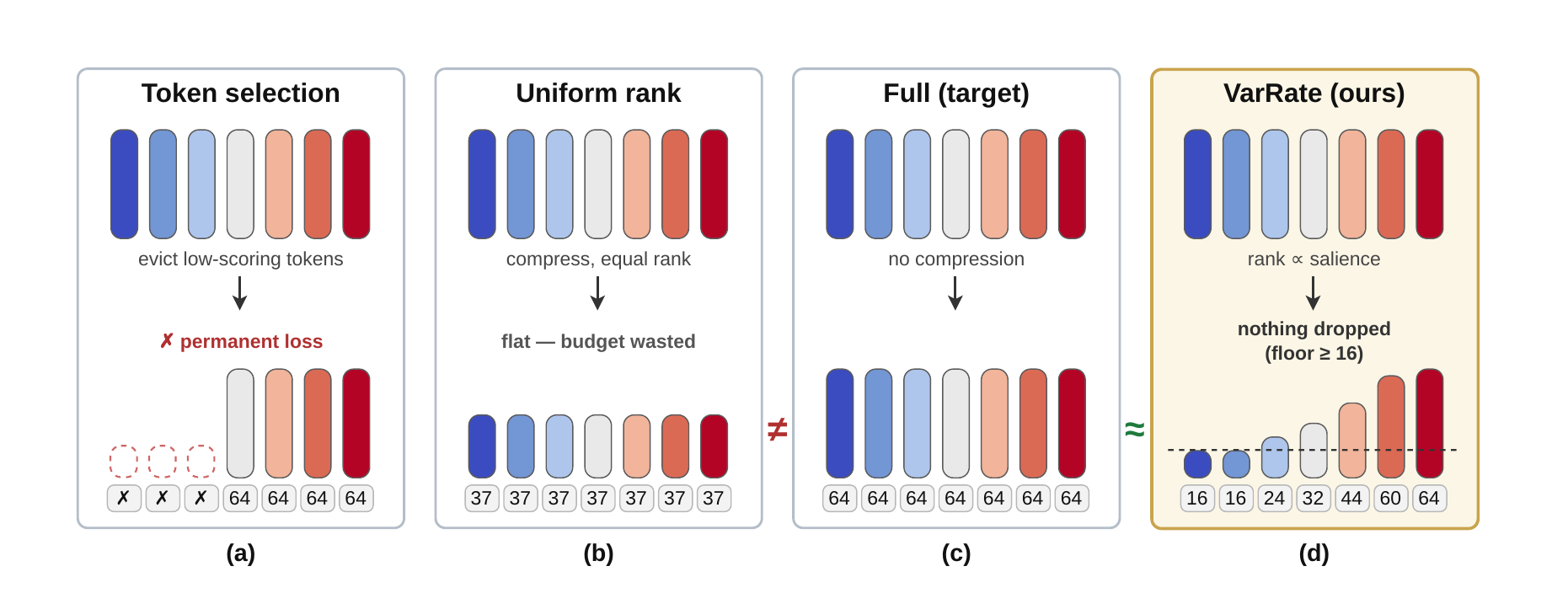}
\caption{\textbf{VarRate allocates rank by salience---it neither evicts tokens nor spreads rank
uniformly.} Each token is a bar whose height is its allocated rank, given below it.
\textbf{(a)}~Token selection evicts the low-scoring tokens; the dropped KV is gone for good, so a
mistaken decision cannot be undone.
\textbf{(b)}~Uniform low-rank coding keeps every token but spends the same rank on all of
them---wasted on the easy tokens, starved on the critical ones.
\textbf{(c)}~The uncompressed cache: every token at full rank, the accuracy target.
\textbf{(d)}~VarRate varies rank with salience and holds every token above a nonzero floor
($r_{\min}\!=\!16$), reaching the target at (a) and (b)'s budget.
Panels (a), (b), (d) use the same KV budget; (c) is uncompressed. Ranks are schematic.}
\label{fig:teaser}
\end{figure*}

\section{Introduction}
\label{sec:intro}

Large language models (LLMs) increasingly operate over long contexts---multi-document
question answering (QA), repository-scale code, and book-length summarization~\cite{bai2024longbench}.
What makes autoregressive inference over such inputs practical is the key--value (KV) cache,
which stores the attention keys and values~\cite{vaswani2017attention} of past tokens so they need
not be recomputed at each step. The cache, however, grows linearly with context length, and at
long context it---not the model's weights---comes to dominate GPU memory at decode
time~\cite{streamingllm}. Compressing the KV cache while preserving accuracy has therefore become a
central problem for efficient LLM serving.

Two training-free families dominate. \emph{Token-selection} methods keep a fixed-size cache and
\emph{evict} the rest (Figure~\ref{fig:teaser}a). The strongest of them---SnapKV~\cite{snapkv} and the
budget-refining successors it inspired~\cite{pyramidkv,adakv}---score each token from a recent
observation window, a signal that is essentially the current query; earlier variants select
positionally or by cumulative attention~\cite{streamingllm,h2o}. Query-aware selection is excellent
when the query is known at compression time. But compress a document once and serve many queries
against it---prefix caching, multi-turn dialogue---and the signal no longer describes what will be
asked; because eviction is irreversible, every token it misjudges is gone for good, and accuracy
collapses. \emph{Uniform low-rank coding}~\cite{palu} takes the
opposite stance: it keeps every token but projects each onto the same reduced-rank subspace
(Figure~\ref{fig:teaser}b). It discards no token, yet spends an identical rank budget on trivial and
on critical tokens---wasting it where it is not needed and starving it where it is.
Quantization~\cite{kivi,qjl}, a third axis, lowers numerical precision and is orthogonal to both.
The middle ground neither family occupies is the natural one: \emph{keep every token, but give each
the rank it deserves}. Methods that do vary rank per token either train a gate for it~\cite{dynakv}
or split it by fixed position~\cite{ojakv}---never from the model's own attention.

A cache fails under reuse only when two conditions coincide: the importance signal has gone stale, and
the method's response to a low score is destructive. Query-aware selection has both. One escape is to
repair the signal---KVzip~\cite{kvzip} scores tokens by how well the cache reconstructs the context, a
query-agnostic criterion, and is robust accordingly---but it pays by re-encoding the entire context
with the full model, several times the cost of the prefill it compresses. We take the other escape and
remove the second condition: \emph{rank should be allocated, not evicted}. A graded, reversible rank
budget keeps every token at some fidelity, so a stale signal merely spends the budget
suboptimally---a misjudged token is coarsened, never discarded---and the signal may then stay cheap,
because being wrong is no longer fatal. We realize this as \textbf{VarRate}, a training-free KV codec that gives each
token a \emph{variable} low-rank budget in proportion to its query salience, via water-filling over a
shared low-rank basis, holding salient tokens at high rank and the rest above a nonzero floor
(Figure~\ref{fig:teaser}d). To our knowledge, VarRate is the first \emph{training-free} method to set the per-token rank from a
query-salience signal---a graded budget, not a fixed positional split.

This design pays off precisely where query-aware selection is weakest. On LongBench across two model
families, at a matched 20\% KV budget VarRate stays within a point of the uncompressed model and
significantly outperforms its own uniform-rank ablation on both. Under query-agnostic reuse,
query-aware selection loses 11--15 points while VarRate loses only 3.5--5.5, and it strictly dominates
the published low-rank codec Palu at every budget. It comes within a point of KVzip's
accuracy---which remains marginally ahead on average---at roughly one-eighth of the compression cost.
We do not claim to beat every baseline---against memory-matched quantization VarRate ties---but it
delivers the robustness of adaptive-rank compression with none of its usual training cost.

\noindent\textbf{Contributions.}
\begin{itemize}
  \item We show that the two dominant training-free families fail for complementary
    reasons---query-aware selection because a stale signal meets an irreversible response, uniform
    coding because it spends rank without regard to salience---and that \emph{allocating} rank by
    salience cures both (Figure~\ref{fig:teaser}).
  \item We present \textbf{VarRate}, the first training-free variable-rate low-rank KV codec:
    per-token rank set by query salience via water-filling over a shared basis, floored so no token
    is dropped, with no training, fine-tuning, or architecture change.
  \item Across two model families and 16 LongBench tasks, VarRate stays within a point of the
    uncompressed model at a 20\% budget, significantly beats uniform-rank coding, strictly dominates
    Palu, and---unlike query-aware selection---stays robust under query-agnostic reuse, coming within
    a point of the purpose-built KVzip at ${\sim}8\times$ lower compression cost.
\end{itemize}

\section{Related Work}
\label{sec:related}

\noindent\textbf{Token selection (eviction).}
The dominant training-free approach keeps a fixed-size cache by scoring tokens and discarding the
rest. StreamingLLM~\cite{streamingllm} retains attention sinks and a recent window; H2O~\cite{h2o}
evicts by cumulative attention; SnapKV~\cite{snapkv} scores importance from a recent observation
window; and PyramidKV~\cite{pyramidkv} and Ada-KV~\cite{adakv} refine the eviction \emph{budget}
across layers and heads; SlimInfer~\cite{sliminfer} prunes tokens dynamically during the forward pass. The accuracy these methods forfeit across long-context tasks has been benchmarked systematically~\cite{kvcompbench}. In the query-aware members---the strongest of them---two weaknesses compound:
the importance signal is essentially the current query, so it goes stale once a cache is reused across
queries, and eviction is \emph{irreversible}, so the tokens that stale signal misjudges cannot be
recovered. VarRate reuses SnapKV's salience signal, but to \emph{allocate} rank rather than to
evict---a misjudged token is coarsened to a low rank, never destroyed.

\noindent\textbf{Low-rank KV compression.}
A second family reduces the \emph{dimension} of each token's key and value rather than the number of
tokens. Palu~\cite{palu} and Eigen Attention~\cite{eigenattn} project the cache onto a fixed low-rank subspace: they discard no token, but
assign every token the same rank. Codecs that \emph{adapt} the rank instead do so mostly across
layers and heads---MatryoshkaKV~\cite{matryoshkakv} tunes this by training, whereas LoRC~\cite{lorc}
sets it post-hoc, layer by layer, without training. Two methods vary rank \emph{per token}: DynaKV~\cite{dynakv} learns a gating network by
fine-tuning, while OjaKV~\cite{ojakv} is training-free but allocates by position---full-rank anchors
and one low rank for the rest, over an online Oja-adapted basis. VarRate fills the remaining cell: to
our knowledge it is the first \emph{training-free} method to set a \emph{graded} per-token rank from a
query-salience signal, by water-filling over a shared basis, with every token held above a nonzero
floor.

\noindent\textbf{Quantization.}
An orthogonal axis lowers numerical precision instead of rank or token count. KIVI~\cite{kivi} uses
per-channel 2-bit keys and per-token values, while KVQuant~\cite{kvquant} and QJL~\cite{qjl} push to
lower bit-widths with reduced overhead. Because it reduces precision rather than rank, quantization is
complementary to VarRate; the two compose, which we confirm empirically.

\noindent\textbf{Query-agnostic reuse and merging.}
When a cache is compressed once and reused across many queries, a query-dependent score fails by
construction. Two methods answer this by making the \emph{signal} query-agnostic: KVzip~\cite{kvzip}
scores tokens by how well the cache reconstructs the context, and Expected
Attention~\cite{expectedattn} estimates future-query attention in closed form. KVzip is genuinely
robust---and it \emph{evicts}, which shows that eviction alone is not the culprit: the culprit is a
stale signal met by an irreversible response. Its price is re-encoding the context with the full
model, several times the prefill it compresses; Expected Attention is far cheaper but does not hold up
in this regime in our evaluation. VarRate removes the irreversible response instead of repairing the
signal, coming within a point of KVzip's accuracy at roughly one-eighth of its cost. A related line \emph{merges} rather than drops evicted entries, as in CaM~\cite{cam}
and KeepKV~\cite{keepkv}, avoiding hard loss but introducing attention-distribution shifts that VarRate's
reconstruction-based coding avoids.

\section{VarRate}
\label{sec:varrate}
\begin{figure*}[t]
\centering
\includegraphics[width=\textwidth]{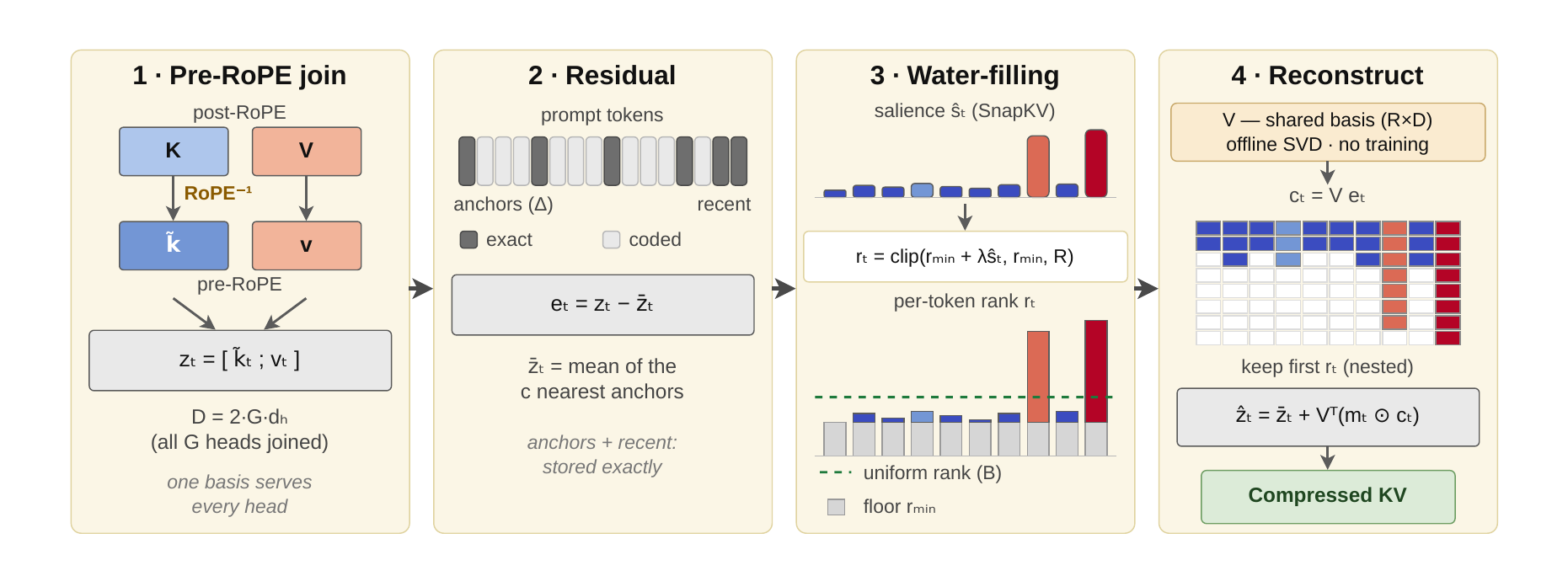}
\caption{\textbf{VarRate, one layer.}
\textbf{(1)}~Keys are un-rotated to pre-RoPE and joined with the values into one
per-token vector $z_t$, so a single basis serves every head.
\textbf{(2)}~Strided anchors and the recent window are stored exactly; every other
token is coded as a residual $e_t$ to the mean of its $c$ nearest anchors.
\textbf{(3)}~SnapKV salience $\hat{s}_t$ is water-filled into a per-token rank $r_t$
at the flat codec's budget. Each bar \emph{is} the allocation: a grey floor
$r_{\min}$ that every token keeps, plus what salience buys above it. Because
$\sum_{t\in\mathcal{C}} r_t = B$, the bar area above the uniform-rank line equals the area below
it---VarRate spends exactly the flat codec's budget, only redistributed.
\textbf{(4)}~A single projection onto the shared basis is truncated per token to its
leading $r_t$ coefficients; every column retains at least $r_{\min}$, so no token is
dropped. The reconstructed keys are finally re-rotated to post-RoPE. Ranks are
schematic.}
\label{fig:pipeline}
\end{figure*}

We present VarRate, a training-free codec that compresses the KV cache by
giving each token a \emph{variable} low-rank budget, allocated by query
salience under a fixed memory budget. The codec is applied independently at
every layer with its own calibrated basis; we describe one layer and drop the
layer index. Figure~\ref{fig:pipeline} gives the pipeline and
Algorithm~\ref{alg:varrate} states it in full.

\subsection{Preliminaries and Codec Setup}
\label{sec:prelim}
\noindent\textbf{Attention and the KV cache.} Each layer has $H$ query heads and
$G\le H$ key--value heads of width $d_h$ (grouped-query attention, GQA). In head $h$, a
query $q$ attends to the cached keys and values $\{(k^h_t,v^h_t)\}$ of the tokens
seen so far and returns $\sum_t \alpha^{h}_{q,t}\,v^h_t$, with attention weights
$\alpha^{h}_{q,t}=\mathrm{softmax}_t(\langle q,k^h_t\rangle/\sqrt{d_h})$. Caching
these keys and values avoids recomputing them at each step, but the cache grows
linearly with the prompt length $L$ and comes to dominate memory at long
context. VarRate compresses it; we first fix its representation.

\noindent\textbf{Why allocate rank.} Token selection and uniform low-rank coding
are two extremes of a single choice---how much of each token to keep. Selection
keeps a few tokens in full and discards the rest; uniform coding keeps every
token at one reduced rank. Neither matches effort to importance. A token's value
enters the attention output weighted by the attention it receives, and an error
in its key perturbs that output largely through the same weight, so a coding
error in a heavily attended token distorts the output far more than the same
error in a diffuse one. VarRate exploits this asymmetry: it spends the rank
budget where attention concentrates, keeping query-salient tokens near-exact and
coding the diffuse remainder coarsely.

\noindent\textbf{Joint per-token vector.} We invert the rotary position embedding
(RoPE), un-rotating each post-RoPE key to its pre-RoPE form, and stack the pre-RoPE
keys and values of all $G$ heads into one per-token
vector:
\begin{equation}
z_t = [\,\tilde{k}_t\,;\, v_t\,]\in\mathbb{R}^{D},\qquad D = 2Gd_h ,
\end{equation}
writing $\tilde{k}_t$ and $v_t$ for the all-head stacks. A single basis per layer
therefore serves all heads jointly. RoPE's per-position rotation inflates the
numerical rank of the key cache, so coding is far more effective before it; prior
low-rank KV compression likewise operates on the pre-RoPE
representation~\cite{palu,deepseekv2}.

\noindent\textbf{Residual codec.} We keep a strided set of anchor tokens
$\mathcal{A}=\{z_1,z_{1+\Delta},z_{1+2\Delta},\dots\}$ exact and code every
other token by its residual to a local anchor mean:
\begin{equation}
\bar{z}_t=\tfrac{1}{c}\!\!\sum_{a\in\mathcal{N}_c(t)}\!\! z_a,\qquad
e_t = z_t-\bar{z}_t,
\end{equation}
where $\mathcal{N}_c(t)$ are the $c$ anchors nearest $z_t$ in Euclidean
distance. Following selection methods we also keep the most recent $w$ tokens
exact. Write $\mathcal{E}$ for this exact set, $\mathcal{C}$
for the coded set, and $n_{\mathrm{ref}}=|\mathcal{E}|$.

\noindent\textbf{Shared basis (training-free).} Offline, on a handful of
unlabeled calibration contexts, we gather the residuals $\{e_t\}$ at each layer
and take the top-$R$ right singular vectors by singular value decomposition (SVD)---a
per-layer principal component analysis (PCA) basis of the residuals---
$V=[\mathbf{v}_1;\dots;\mathbf{v}_R]\in\mathbb{R}^{R\times D}$, ordered by
decreasing singular value. This needs no gradients and no fine-tuning; VarRate
runs on off-the-shelf models. The basis is a model-side, one-time cost shared by
every sequence---one $R\times D$ matrix per layer, $134$\,MB across all layers in half precision
for Llama-3.1-8B---and, following the convention of prior low-rank
codecs~\cite{palu}, it is not charged to the per-sequence cache budget.

\noindent\textbf{Budget.} Let $\kappa\in(0,1)$ be the target KV budget (the
fraction of the full cache retained). The exact set costs $n_{\mathrm{ref}}D$
scalars, and a token coded at rank $r$ costs $r$ coefficients. Matching total
storage to $\kappa L D$ leaves a coded budget
\begin{equation}
B=\kappa L D-n_{\mathrm{ref}}D=D\,(\kappa L-n_{\mathrm{ref}}),
\label{eq:budget}
\end{equation}
to be distributed as ranks over the coded tokens,
$\sum_{t\in\mathcal{C}}r_t=B$. Each coded token additionally stores the indices
of its $c$ anchors, which we count in the reported footprint; they add $0.2\%$
of the cache. A uniform-rank codec spends the budget as a constant
$r_t=B/|\mathcal{C}|$ for every token; everything to this point---the reference
codec and the shared basis---is common to that baseline, and VarRate's
contribution is how it sets $r_t$ instead.

\subsection{Salience-Guided Variable-Rate Allocation}
\label{sec:alloc}
\noindent\textbf{Salience.} We score each token by the attention it draws from a
recent observation window---the signal introduced by SnapKV~\cite{snapkv}. Writing
$\mathcal{W}$ for the queries of the last $w_{\mathrm{obs}}$ tokens, we average the
attention weights $\alpha^{h}_{q,t}$ that this window places on token $t$,
\begin{equation}
s_t=\frac{1}{|\mathcal{W}|\,H}\sum_{q\in\mathcal{W}}\ \sum_{h=1}^{H}\alpha^{h}_{q,t},
\qquad t\in\mathcal{C},
\label{eq:salience}
\end{equation}
apply SnapKV's length-$5$ moving average over adjacent tokens (its clustering
step), and min--max normalize over the coded set to $\hat{s}_t\in[0,1]$. The window
is simply the tail of the prompt: when the prompt carries the question---the
standard setting---this is a sharp signal, but when a document is compressed
\emph{once} and reused across many queries, no question is present and $\hat{s}$
is only a stale proxy for what will be asked. VarRate does not try to repair the
signal; it makes a stale one survivable, coarsening a misjudged token to
a low rank rather than deleting it.

\noindent\textbf{Water-filling.} We allocate $B$ across coded tokens in
proportion to salience, above a floor $r_{\min}$ and capped at the basis rank
$R$:
\begin{equation}
r_t=\mathrm{clip}\!\big(r_{\min}+\lambda\,\hat{s}_t,\; r_{\min},\; R\big),
\qquad t\in\mathcal{C},
\label{eq:waterfill}
\end{equation}
where the single multiplier $\lambda>0$ is set so that
$\sum_{t\in\mathcal{C}}r_t=B$. This is the standard water-filling recursion
(Algorithm~\ref{alg:varrate}): raise the ``water level'' $\lambda$, clip any
token that reaches $R$, and redistribute its surplus, until the budget is
met.\footnote{Ranks are integers in the implementation, so realized storage
exceeds $B$ by under $0.4\%$ (measured). The uniform-rank codec rounds
identically, so matched-memory comparisons are unaffected.} Two properties hold
\emph{by construction}: (i) total storage equals the uniform-rank codec's at the
same $\kappa$, so every comparison is at matched memory; and (ii) since
$r_{\min}>0$, every token keeps at least rank $r_{\min}$---\emph{no token is ever
dropped}, unlike eviction. The two familiar regimes are its limits: an
uninformative signal (constant $\hat{s}$) collapses Eq.~\ref{eq:waterfill} to the
uniform-rank codec, $r_t=B/|\mathcal{C}|$, while eviction is the degenerate
$r_t=0$ that the floor forbids. VarRate lies strictly between them---graded
ranks, none of them zero.

\begin{algorithm}[t]
\caption{VarRate compression (one layer, prefill)}
\label{alg:varrate}
\textbf{Input}: post-RoPE keys/values $\{(k_t,v_t)\}_{t=1}^{L}$; basis
$V\in\mathbb{R}^{R\times D}$; budget $\kappa$; floor $r_{\min}$; stride $\Delta$;
windows $w$, $w_{\mathrm{obs}}$\\
\textbf{Output}: compressed keys/values
\begin{algorithmic}[1]
\STATE un-rotate keys to pre-RoPE;\;\; $z_t\leftarrow[\,\tilde{k}_t\,;\,v_t\,]$
\STATE $\mathcal{A}\!\leftarrow\!$ tokens at stride $\Delta$;\; $\mathcal{E}\!\leftarrow\!\mathcal{A}\cup\{\text{last }w\text{ tokens}\}$;\; $\mathcal{C}\!\leftarrow\![L]\setminus\mathcal{E}$
\STATE $\bar{z}_t\!\leftarrow\!$ mean of the $c$ nearest anchors;\;\; $e_t\!\leftarrow\! z_t-\bar{z}_t$
\STATE $s_t\!\leftarrow\!$ SnapKV score over a length-$w_{\mathrm{obs}}$ window;\;\; $\hat{s}_t\!\leftarrow\!$ min--max over $\mathcal{C}$
\STATE $B\leftarrow D(\kappa L-|\mathcal{E}|)$
\STATE $r_t\leftarrow r_{\min}\ (t\in\mathcal{C})$;\;\; $\mathcal{U}\leftarrow\mathcal{C}$ \COMMENT{water-filling}
\REPEAT
\STATE $\delta\leftarrow B-\sum_{u\in\mathcal{C}}r_u$
\STATE $r_t\leftarrow r_t+\delta\,\hat{s}_t\big/\!\sum_{u\in\mathcal{U}}\hat{s}_u\ \ (t\in\mathcal{U})$
\STATE $\mathcal{S}\leftarrow\{t\in\mathcal{U}: r_t>R\}$;\;\; $r_t\leftarrow R\ (t\in\mathcal{S})$;\;\; $\mathcal{U}\leftarrow\mathcal{U}\setminus\mathcal{S}$
\UNTIL{$\mathcal{S}=\emptyset$}
\STATE $\mathbf{c}_t\!\leftarrow\! Ve_t$;\;\; $\hat{z}_t\!\leftarrow\!\bar{z}_t+V^{\!\top}(\mathbf{1}[j\!\le\! r_t]\odot\mathbf{c}_t)\ (t\in\mathcal{C})$
\STATE $\hat{z}_t\leftarrow z_t\ (t\in\mathcal{E})$
\STATE split $\hat{z}_t$ into key and value; re-rotate the key to post-RoPE
\STATE \textbf{return} compressed keys/values
\end{algorithmic}
\end{algorithm}

\subsection{Reconstruction and Complexity}
\label{sec:cost}
Each coded token is projected once onto the shared basis, then \emph{truncated}
to its rank. With coefficients $\mathbf{c}_t=Ve_t\in\mathbb{R}^{R}$ and a nested
mask $m_{t,j}=\mathbf{1}[\,j\le r_t\,]$,
\begin{equation}
\hat{z}_t=\bar{z}_t+\!\!\sum_{j\le r_t}\!\! c_{t,j}\,\mathbf{v}_j
=\bar{z}_t+V^{\!\top}(m_t\odot\mathbf{c}_t),\quad t\in\mathcal{C},
\label{eq:recon}
\end{equation}
and $\hat{z}_t=z_t$ for $t\in\mathcal{E}$. Because $V$ is ordered by singular
value, its leading directions carry the most residual energy on the calibration
distribution; and because $V$ is orthonormal, enlarging $r_t$ monotonically
reduces the reconstruction error of the \emph{same} projection---a nested,
Matryoshka-style truncation obtained without any training. Finally $\hat{z}_t$
is split into key and value and the key is re-rotated to post-RoPE.

\noindent\textbf{Cost.} The projection $Ve_t$ is computed once per token at the
full rank $R$ and then masked, so the variable rate adds no arithmetic over a
fixed-rank codec; its only overhead is keeping all $R$ rows of the basis
resident, where a flat code needs only the $r$ it uses. Beyond this the codec
adds, per layer, a nearest-anchor search in $O(L^2 d_h G/\Delta)$ and two
rank-$R$ projections in $O(LDR)$, all inside the single prefill pass the model
already runs---it never re-encodes the context to score tokens. Both are far
smaller than the prefill's own attention, and Section~\ref{sec:experiments}
measures the resulting overhead.

\noindent\textbf{Composability.} The coefficients $\mathbf{c}_t$ are themselves
quantizable; standard group-wise $b$-bit quantization of them composes VarRate
with the orthogonal quantization axis, leaving the allocation unchanged.

\section{Experiments}
\label{sec:experiments}

\subsection{Setup}
\noindent\textbf{Models and benchmark.} We evaluate on Llama-3.1-8B-Instruct and
Qwen2.5-7B-Instruct, and cross-check on Mistral-7B-Instruct-v0.2. All results are on
LongBench~\cite{bai2024longbench}: 16 tasks under their official metrics, spanning
single- and multi-document QA, summarization, few-shot learning, synthetic retrieval
and code completion, with $n{=}200$ examples per task and greedy decoding. Confidence
intervals are taken over examples, so no seed averaging is required. Comparisons
between methods are paired over the 16 tasks; we report a 20{,}000-sample paired
bootstrap 95\% confidence interval (CI) and an exact sign test, and count a claim as supported only when the
CI excludes zero.

\noindent\textbf{Implementation.} VarRate is implemented in KVPress. Unless stated
otherwise we compress to a KV budget of $\kappa{=}0.20$ (a $5\times$ reduction), with
floor $r_{\min}{=}16$, basis rank $R{=}1024$, stride $\Delta{=}16$, $c{=}4$ nearest
anchors, and windows $w{=}w_{\mathrm{obs}}{=}64$. Each layer's basis is calibrated once, offline, from six
unlabeled LongBench contexts---no gradients, no labels, no fine-tuning---and reused
for every task. Accuracy is measured on a single A100.

\noindent\textbf{Baselines.} We compare against the uncompressed cache and against one
compression axis at a time: token selection---SnapKV~\cite{snapkv},
PyramidKV~\cite{pyramidkv}, and Ada-KV~\cite{adakv}; low-rank coding---Palu~\cite{palu}
and our own uniform-rank ablation, ``flat''; quantization---KIVI~\cite{kivi} at 2 and
4 bits; and the two methods built for query-agnostic reuse, KVzip~\cite{kvzip} and
Expected Attention~\cite{expectedattn}. All run from one KVPress installation at a
matched budget, and every footprint we report is \emph{measured} rather than nominal
(Table~\ref{tab:cost}).

\begin{table*}[t]
\centering\small
\setlength{\tabcolsep}{5pt}
\begin{tabular}{lcccccccc}
\toprule
method & mem & Single-Doc & Multi-Doc & Summ. & Few-shot & Synthetic & Code & \textbf{AVG-16} \\
\midrule
\multicolumn{9}{l}{\textit{Llama-3.1-8B-Instruct}}\\
Full (ceiling) & 100\% & 44.1 & 47.4 & 29.2 & 54.2 & 55.4 & 50.4 & 46.01 \\
PyramidKV & 20\% & 43.7 & 45.3 & 25.0 & 59.2 & 56.0 & 52.5 & \textbf{46.02} \\
KIVI-4$^{\dagger}$ & 31\% & 44.9 & 47.3 & 29.2 & 53.4 & 56.1 & 49.3 & 45.94 \\
SnapKV & 20\% & 42.9 & 46.7 & 25.3 & 57.7 & 56.3 & 51.1 & 45.79 \\
\textbf{VarRate (ours)} & 20\% & 45.4 & 46.4 & 26.7 & 55.7 & 54.5 & 49.8 & 45.69 \\
Ada-KV & 20\% & 42.6 & 46.8 & 25.5 & 57.2 & 56.0 & 48.5 & 45.33 \\
KIVI-2 & 19\% & 44.7 & 45.3 & 29.1 & 51.9 & 55.1 & 47.0 & 44.82 \\
flat (uniform rank) & 20\% & 41.3 & 45.1 & 26.5 & 52.6 & 48.4 & 51.1 & 43.47 \\
\midrule
\multicolumn{9}{l}{\textit{Qwen2.5-7B-Instruct}}\\
Full (ceiling) & 100\% & 41.9 & 41.6 & 27.8 & 66.3 & 56.0 & 67.9 & 48.79 \\
KIVI-4$^{\dagger}$ & 31\% & 42.6 & 43.0 & 27.8 & 65.5 & 56.2 & 67.2 & 48.98 \\
\textbf{VarRate (ours)} & 20\% & 40.7 & 41.9 & 25.8 & 65.8 & 56.0 & 66.7 & \textbf{48.00} \\
KIVI-2 & 19\% & 40.0 & 41.9 & 27.6 & 64.5 & 53.2 & 60.2 & 46.82 \\
Ada-KV & 20\% & 39.2 & 39.6 & 24.6 & 62.6 & 56.0 & 67.0 & 46.49 \\
SnapKV & 20\% & 40.0 & 39.4 & 24.6 & 61.1 & 55.5 & 66.8 & 46.25 \\
PyramidKV & 20\% & 38.0 & 39.0 & 23.3 & 58.5 & 55.2 & 65.8 & 44.90 \\
flat (uniform rank) & 20\% & 31.4 & 13.9 & 24.8 & 49.7 & 39.9 & 24.5 & 30.50 \\
\bottomrule
\end{tabular}
\caption{LongBench (16 tasks) at a matched $20\%$ KV budget ($5\times$ compression).
Bold marks the best \emph{matched-memory} compressor on each model. $^{\dagger}$KIVI-4 stores $31\%$ of the cache ($1.55\times$ the
others) and is therefore not a matched comparison. Two-model means at matched memory:
\textbf{VarRate 46.85}, SnapKV 46.02, Ada-KV 45.91, KIVI-2 45.82, PyramidKV 45.46,
flat 36.99.}
\label{tab:main}
\end{table*}

\begin{figure*}[t]
\centering
\includegraphics[width=\textwidth]{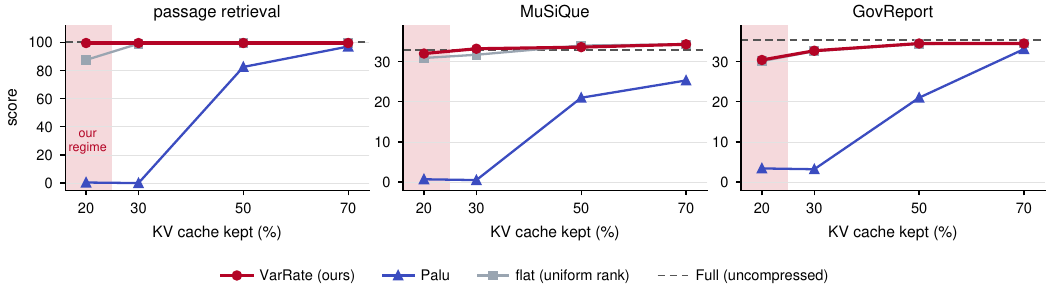}
\caption{Budget sweep against the published low-rank codec (Llama-3.1-8B). Palu was
built for the keep-$50$--$70\%$ regime: in ours it is at essentially zero, and only
becomes viable at keep-$50\%$. The sweep also shows, honestly, where VarRate's margin
over uniform rank lives---largest where the budget is tightest, closing as it loosens.}
\label{fig:budget}
\end{figure*}

\subsection{Accuracy at a Matched Budget}
At $5\times$ compression VarRate lands within $0.3$ points of the uncompressed ceiling
on Llama and $0.8$ on Qwen (Table~\ref{tab:main}), and it has the best two-model mean
of any matched-memory compressor.

The per-model picture is more nuanced than that average. On Llama VarRate is
\emph{fourth of seven} compressors: PyramidKV ($46.02$) and
SnapKV ($45.79$) sit nominally above it. Paired over the 16 tasks, however, those gaps
are ties ($-0.33$ and $-0.10$; both CIs straddle zero). On Qwen the ordering reverses
and the gaps become real: VarRate is first, beating PyramidKV by $+3.10$ ($p{=}0.001$)
and SnapKV by $+1.75$ ($p{=}0.022$). Against memory-matched quantization it \emph{ties}
KIVI-2 on both models ($+0.86$ and $+1.18$, both n.s.)---not a win. KIVI-4 does edge it
on Qwen, but stores $31\%$ of the cache, $1.55\times$ what every other method spends.
The defensible cross-model claim is therefore \emph{consistency}: VarRate is never
significantly worse than any matched-memory baseline on either model, and significantly
better than several on Qwen.

\noindent\textbf{What the allocation buys.} The comparison that isolates our
contribution is against our own uniform-rank ablation, and it is the strongest result
we have: $+2.22$ on Llama ($p{=}0.021$, winning 13 of 16 tasks) and $+17.50$ on Qwen
($p{=}0.001$, 15 of 16). Qwen's four-KV-head cache is especially hard for uniform low-rank
coding---flat collapses to $30.50$, losing two thirds of its multi-document QA
score---and spending the same budget by salience recovers almost all of it.

\noindent\textbf{Against the published low-rank codec.} Palu was designed for the
keep-$50$--$70\%$ regime, whereas VarRate operates at keep-$15$--$25\%$; sweeping the
budget (Figure~\ref{fig:budget}) makes the gap stark. At a $20\%$ budget Palu scores
$0.4$ on passage retrieval, $0.7$ on MuSiQue and $3.4$ on GovReport, against VarRate's
$99.5$, $32.0$ and $30.4$; only by keep-$50\%$ does it become viable. Palu's released
checkpoint stores $20.41\%$ of the cache against our measured $20.22\%$, so the
comparison is memory-matched. The same sweep shows where VarRate's margin over uniform
rank lives: it is largest exactly where the budget is tightest, and closes as the
budget loosens.

\begin{table}[tb]
\centering\small
\begin{tabular}{lccc}
\toprule
method & Llama & Qwen & mean \\
\midrule
Full (control) & 59.91 & 58.21 & 59.06 \\
\midrule
KVzip (pinned) & 55.34 & \textbf{56.42} & \textbf{55.88} \\
\textbf{VarRate (ours)} & \textbf{56.21} & 51.79 & 54.00 \\
Ada-KV & 47.48 & 40.28 & 43.88 \\
SnapKV & 45.75 & 41.35 & 43.55 \\
flat (uniform rank) & 42.18 & 22.33 & 32.26 \\
Expected Attention & 32.83 & 26.14 & 29.49 \\
\bottomrule
\end{tabular}
\caption{Query-agnostic reuse: the document is compressed once, with the question
absent from the compressed context. Three-task mean (passage / MuSiQue / Qasper).
KVzip is given the same 64-token pin as VarRate. The uncompressed control barely moves,
so the collapses are caused by the compression.}
\label{tab:agnostic}
\end{table}

\subsection{Query-Agnostic Reuse}
Compressing a document once and serving many queries against it removes the question
from the compressed context, and with it the signal every query-aware method depends
on. The uncompressed control barely moves ($+0.08$ and $+0.55$), so what follows is
caused by the compression and not by the change of prompt.

Selection collapses: SnapKV loses $12.96$ and $14.66$ points and Ada-KV $11.01$ and
$15.28$ (Table~\ref{tab:agnostic}). Expected Attention, though built to be
query-agnostic, collapses hardest of all ($-21.68$)---its estimate of the future query
is drawn from a prefill that ordinarily contains the question. VarRate degrades
gracefully instead, by $3.53$ and $5.50$. Allocation rather than eviction is the
reason: a token that the stale signal misjudges is coarsened to a low rank, never
removed.

\noindent\textbf{A correction we owe KVzip.} When the query is known at compression
time it is appended \emph{into} the compressed context. SnapKV pins that region so it
can never be evicted, and VarRate inherits the pin because it scores with SnapKV;
stock KVzip protects only its first four tokens, and can therefore evict the very
question it is about to be asked. We grant KVzip the identical 64-token pin. Its
budget is unchanged by construction---the pin decides \emph{which} entries are
dropped, never \emph{how many}---and it lifts KVzip by $+0.89$ on Llama and $+3.12$ on
Qwen, erasing an apparent lead of ours. Every KVzip number below is the pinned,
stronger one.

Against KVzip, the method built for this regime, the outcome is genuinely mixed.
Given the same pin, the two
are accuracy-equivalent in three of the four model$\times$regime cells; the fourth---%
Qwen under reuse---goes to \emph{KVzip} by $4.63$ points, and KVzip leads the four-cell
mean by $0.77$. \textbf{We claim no accuracy advantage over KVzip anywhere.} What we
claim is that VarRate reaches comparable robustness by keeping a cheap query-aware
signal and making it survivable, rather than by paying to replace it.

\begin{table}[tb]
\centering\small
\begin{tabular}{lccc}
\toprule
method & KV & prefill & decode \\
\midrule
Full (uncompressed) & 100\% & --- & 37.5 \\
SnapKV & 20.0\% & $+2.0\%$ & 44.4 \\
PyramidKV & 20.0\% & $+2.1\%$ & 44.3 \\
KIVI-2 & 19.6\% & $+5.8\%$ & 37.5 \\
Ada-KV & 19.9\% & $+6.3\%$ & 22.6$^{\ddagger}$ \\
Expected Attention & 20.0\% & $+7.3\%$ & 44.5 \\
flat (uniform rank) & 20.2\% & $+24.4\%$ & 37.4 \\
\textbf{VarRate (ours)} & 20.2\% & $\mathbf{+44.0\%}$ & 37.4 \\
\quad $+$ 4-bit coefficients & 11.0\% & $+45.9\%$ & 37.5 \\
KVzip & 19.9\% & $\mathbf{+372.2\%}$ & 18.8$^{\ddagger}$ \\
\bottomrule
\end{tabular}
\caption{Measured cost, one instrument (Llama-3.1-8B, A100, 16k context). KV is the
measured cache footprint, prefill the overhead over the uncompressed model, decode in
tok/s. $^{\ddagger}$The maskers' decode rate is a harness artifact---KVPress simulates
eviction rather than physically pruning---not a property of those methods.}
\label{tab:cost}
\end{table}

\begin{figure}[tb]
\centering
\includegraphics[width=\columnwidth]{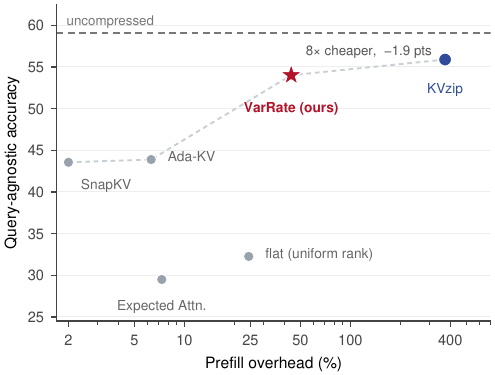}
\caption{Accuracy under query-agnostic reuse against compression cost. Both axes are
measured and the cost axis is logarithmic. The dashed guide traces the trade-off
frontier: VarRate lies on it, coming within $1.9$ points of KVzip at roughly
one-eighth of the prefill overhead. Uniform rank and Expected Attention are
\emph{dominated}---cheaper methods reach higher accuracy than either.}
\label{fig:pareto}
\end{figure}

\subsection{Cost}
VarRate's prefill overhead is $+44.0\%$ (Table~\ref{tab:cost}), against $+2.0\%$ for
selection and $+24.4\%$ for the flat codec. Allocation is not free. The underlying
\emph{arithmetic}, however, is nearly free: the codec's
entire floating-point operation (FLOP) count is $2.61\%$ of one forward pass, so what we measure is unfused
\texttt{fp32} implementation rather than a floor---a bf16 variant cuts the overhead by
$40\%$ with accuracy unchanged (max $|\Delta|{=}0.07$). The contrast with KVzip
is of a different kind:
its $+372.2\%$ is \emph{structural}, since it re-encodes the whole context with the
full model in order to score it. Figure~\ref{fig:pareto} places the two. VarRate comes
within $1.9$ points of KVzip's query-agnostic accuracy at roughly one-eighth of its
cost, while the methods that are genuinely cheap sit at least ten points below
both.

Decoding speed is unchanged ($37.4$ tok/s, against the uncompressed $37.5$),
because the codec
reconstructs a full-length cache; realizing the saving at run time requires a fused
kernel, as Palu's does. We therefore measure accuracy in the standard
reconstruct-in-place mode and cost on a separate instrument, following the convention
of this literature.

\begin{table}[tb]
\centering\small
\begin{tabular}{lcc}
\toprule
configuration & passage & GovReport \\
\midrule
flat (uniform rank) & 87.5 & 30.1 \\
random signal $+$ water-filling & 93.0 & 29.6 \\
salience $+$ binary top-$K$ split & \textbf{99.5} & 29.0 \\
\textbf{salience $+$ water-filling (ours)} & \textbf{99.5} & \textbf{30.4} \\
\bottomrule
\end{tabular}
\caption{Both halves of the method matter (Llama, $20\%$ budget). The \emph{signal}
buys retrieval ($87.5\!\rightarrow\!99.5$); the \emph{graded} allocation is what
preserves summarization---a binary top-$K$ split ties on retrieval but falls to $29.0$,
below even uniform rank.}
\label{tab:ablation}
\end{table}

\subsection{Ablations}
\noindent\textbf{The signal, and the shape of the allocation.} Two things could be
doing the work: the salience signal, or the graded way the budget is spent. Both do
(Table~\ref{tab:ablation}). Replacing SnapKV salience with a random signal costs $6.5$
points on retrieval ($99.5\!\rightarrow\!93.0$), and uniform rank costs $12$
($87.5$)---the signal matters. Replacing water-filling with a binary top-$K$ split,
winners at $R$ and everyone else at the floor, ties on retrieval ($99.5$) but
\emph{loses} on summarization ($29.0$, below even flat's $30.1$): starving the
unselected tokens to the floor destroys exactly the distributed context that low-rank
coding is good at. The graded allocation, not merely the signal, is what lets VarRate
hold coding's summarization strength while gaining selection's retrieval strength.

\noindent\textbf{How good is the signal?} On passage retrieval, an oracle that
allocates rank by the gold answer span reaches $100.0$ and attention salience reaches
$99.0$, against $93.0$ for uniform rank (a separate $n{=}100$ harness). The signal we
can actually compute is within a point of optimal.

\noindent\textbf{Composability.} Quantizing the coefficients to 3 bits leaves accuracy
essentially unchanged ($99.5/31.3/30.4$ against $99.5/32.0/30.4$ on
passage/MuSiQue/GovReport) at $9\%$ of the cache: low-rank coding and quantization
compose, as the method predicts.

\noindent\textbf{Calibration, and a third model.} Rebuilding the basis from different
examples, or from different tasks, moves every result by at most $1.8$ points, inside
the confidence intervals---the codec is not tuned to its calibration set. The pattern
also carries to a third family: on Mistral-7B, uniform rank collapses on retrieval
($29.1$ at a $15\%$ budget) while VarRate holds $73.6$, within two points of the
uncompressed $75.6$.

\noindent\textbf{Robustness checks (signal, floor, budget, long context).} A further
set of controls, provided in the supplementary material, confirms the signal is
content-driven rather than merely graded, replicates both the signal and the allocation
choice on a second model, shows the rank floor is slack rather than a lever, states
VarRate's viable operating range at aggressive budgets, and extends the accuracy result
to $128$K context.

\section{Conclusion}
We asked what a KV codec should do when its importance signal is wrong. Token selection
answers by deleting, and pays for it: under cache reuse, where the signal goes stale, it
loses $11$--$15$ points. VarRate answers by \emph{coarsening} instead---allocating a
variable low-rank budget by query salience over a shared basis, holding every token above
a nonzero floor, and requiring no training at all. The allocation is what carries the
result: it beats its own uniform-rank ablation by $2.22$ and $17.50$ points on two model
families, and the signal it uses lands within a point of an oracle. Against KVzip, a
method purpose-built for the reuse regime, VarRate is accuracy-equivalent in three of
four settings and behind in the fourth---but it gets there at roughly one-eighth of the
cost. Robustness under reuse, we conclude, can be bought by making a cheap signal
survivable rather than by paying to replace it.

\section*{Acknowledgments}
This work used NCSA Delta GPU at NCSA through allocation CIS240646 from the Advanced
Cyberinfrastructure Coordination Ecosystem: Services \& Support (ACCESS)
program~\cite{access}, which is supported by National Science Foundation grants
\#2138259, \#2138286, \#2138307, \#2137603, and \#2138296.

\bibliography{varrate}

\clearpage
\appendix

\section{Additional Experiments and Analyses}

\subsection{Perplexity at Equal Budget}
\label{app:ppl}
Every other result in this paper is a downstream task metric (exact match, ROUGE,
F1), which conflates the quality of the compressed representation with the
model's ability to work around a degraded one. As an independent sanity check, we
measure next-token perplexity on WikiText-2~\cite{merity2016pointer}, a plain
causal-language-modeling corpus unrelated to LongBench, with VarRate and the flat
codec compressed to exactly the same storage budget (not merely the same nominal
$\kappa$, but the same realized byte count, so the comparison cannot be won by
either method rounding differently). VarRate reaches $6.95$ PPL against flat's
$7.26$---a $0.31$-point improvement in raw predictive likelihood, not a
task-specific score. Because perplexity has no notion of ``retrieval'' or
``summarization,'' this isolates the claim to its purest form: salience-weighted
rank allocation reconstructs the underlying key--value geometry more faithfully
than uniform rank allocation, at identical cost, independent of what downstream
task later reads that cache.

\subsection{VarRate Against the Selection Family Across Budgets}
\label{app:selfam}
Table~\ref{tab:selfam} sweeps SnapKV, PyramidKV, and Ada-KV against VarRate at
keep-$15/20/25\%$ on Llama. VarRate ties the family on retrieval and multi-hop QA
(differences inside CIs at every budget) and wins summarization outright, by
$1.0$--$2.8$ points over the best selection method---significant at keep-$20/25\%$.
This is a Pareto improvement over selection specifically: VarRate matches it where it
is strong and wins where uniform attention-window scoring is structurally weak
(distributed, low-salience summarization context). The pattern holds at every budget
in the sweep, not just at the $20\%$ operating point used in the main text: the
summarization margin is present already at keep-$15\%$ ($28.6$ vs.\ the best
selection method's $27.8$) and widens to $32.2$ vs.\ $29.5$ at keep-$25\%$, while
retrieval and multi-hop stay statistically tied throughout. This matters because it
rules out an alternative explanation for Table~\ref{tab:main}'s summarization
result---that it is a one-off artifact of the specific $20\%$ budget chosen for the
headline comparison. Three independent selection methods (SnapKV, PyramidKV, Ada-KV)
share the same observation-window scoring mechanism and inherit the same structural
weakness on distributed, low-salience context; VarRate's advantage tracks that shared
weakness rather than any one competitor's idiosyncrasies, which is why it appears
consistently across the whole family and across the whole budget range.

\begin{table}[!ht]
\centering\small
\begin{adjustbox}{max width=\columnwidth}
\begin{tabular}{lccc}
\toprule
method (KR15/20/25) & passage & musique & gov\_report \\
\midrule
SnapKV & 99.5/100/100 & 30.9/32.2/32.7 & 27.8/28.8/29.4 \\
PyramidKV & 99.5/100/100 & 27.8/27.7/28.8 & 27.0/28.2/29.2 \\
Ada-KV & 100/100/100 & 31.7/32.6/32.5 & 27.6/28.8/29.5 \\
\textbf{VarRate} & 99.5/99.5/99.5 & 30.3/32.0/33.2 & \textbf{28.6/30.4/32.2} \\
\bottomrule
\end{tabular}
\end{adjustbox}
\caption{Selection-family budget sweep, Llama-3.1-8B. VarRate ties on retrieval and
multi-hop QA at every budget and leads summarization at every budget.}
\label{tab:selfam}
\end{table}

\subsection{Quantization: Where It Wins, and Where It Composes}
\label{app:kivi}
KIVI (2- and 4-bit) is a strong orthogonal axis, not a method VarRate dominates.
Reproduced in-pipeline at effective memory $\approx19\%$ (KIVI-2) and $\approx31\%$
(KIVI-4): on Llama, KIVI reaches $35.2$--$35.5$ on GovReport against VarRate's
$30.4$, because quantization preserves \emph{distributed} information across all
tokens at low precision, while rank reduction discards information outright. VarRate
ties or edges KIVI on retrieval and multi-hop QA. The mechanism behind the split is
worth stating explicitly: quantization keeps every token at full rank and only
coarsens the numerical precision of each coordinate, so information that is spread
thinly across many tokens---exactly what a long, diffuse document like GovReport
demands---survives at low bit-width. Low-rank coding does the opposite: it keeps
full numerical precision but discards entire coordinate directions, which is fatal
precisely for that same distributed information, no matter how the discarded rank
is allocated across tokens. This is a property of the compression \emph{axis}, not
of any one implementation, so we do not expect a better salience signal or a
different water-filling schedule to close the gap. The honest claim is therefore
``the best low-rank codec, competitive with and composable with quantization''---not
a universal win, and Table~\ref{tab:main}/\ref{tab:cost} already show the two compose
cleanly (VarRate$+$3-bit coefficients: $9\%$ of the cache, accuracy within noise of
fp16 VarRate on all three tasks). That composability is the practical resolution:
rather than choosing between the two axes, a deployment can quantize VarRate's
coefficients and recover quantization's footprint while keeping the allocation
mechanism's retrieval and multi-hop strength.

\begin{table}[!ht]
\centering\small
\begin{adjustbox}{max width=\columnwidth}
\begin{tabular}{llccc}
\toprule
model & method & passage & musique & gov \\
\midrule
Llama & KIVI-2 & 100.0 & 30.2 & \textbf{35.2} \\
Llama & KIVI-4 & 99.5 & 32.7 & \textbf{35.5} \\
Llama & \textbf{VarRate} & 99.5 & 32.0 & 30.4 \\
Qwen & KIVI-2 & 95.0 & 30.8 & \textbf{33.3} \\
Qwen & KIVI-4 & 99.5 & 29.2 & \textbf{33.8} \\
Qwen & \textbf{VarRate} & 99.5 & 28.7 & 31.3 \\
\bottomrule
\end{tabular}
\end{adjustbox}
\caption{VarRate vs.\ the quantization axis (in-pipeline KIVI reproduction). KIVI
wins summarization (gov\_report) on both models; VarRate ties or edges it on
retrieval/multi-hop.}
\label{tab:kivi}
\end{table}

\subsection{Composability at Multiple Bit-Widths}
\label{app:compose}
The main paper reports that quantizing VarRate's low-rank coefficients to 3 bits
leaves accuracy unchanged at $9\%$ of the cache; Table~\ref{tab:compose} shows the
full sweep behind that single number. Accuracy degrades smoothly and only slightly as
bit-width drops from fp16 through 8-, 4-, and 3-bit coefficients, and even at 3 bits
every task is within noise of the fp16 codec (passage exact, musique $-0.7$, gov
exact), while the footprint more than halves ($20\%\to9\%$). This is the same
composability pattern the published low-rank literature reports for coefficient
quantization on top of a fixed-rank codec (Palu$+$quantization); the fact that it
holds here too, on a codec whose rank is itself variable per token, indicates the two
compression axes are genuinely orthogonal rather than coincidentally compatible at
one specific setting.

\begin{table}[tb]
\centering\small
\begin{adjustbox}{max width=\columnwidth}
\begin{tabular}{lcccc}
\toprule
config (Llama, KR20) & passage & musique & gov & memory \\
\midrule
VarRate fp16 & 99.5 & 32.0 & 30.4 & $\sim$20\% \\
VarRate + 8-bit & 99.5 & 31.8 & 30.3 & $\sim$14\% \\
VarRate + 4-bit & 99.5 & 31.9 & 30.4 & $\sim$10\% \\
\textbf{VarRate + 3-bit} & \textbf{99.5} & \textbf{31.3} & \textbf{30.4} & $\mathbf{\sim}$\textbf{9\%} \\
\bottomrule
\end{tabular}
\end{adjustbox}
\caption{Coefficient quantization sweep. Accuracy is within noise of fp16 all the way
down to 3-bit coefficients, at less than half the memory.}
\label{tab:compose}
\end{table}

\subsection{Additional Baselines: Sanity Checks Behind the Comparisons}
\label{app:baselines}
Two validation steps underlie the baselines used throughout this paper and the main
text, neither of which is visible in the headline numbers themselves. First, KIVI
(2- and 4-bit) is reproduced in-pipeline rather than taken from published numbers;
before trusting its comparison, we validate the reproduction directly on
uncompressed-equivalent passage retrieval, where KIVI-2 ($\approx19\%$ memory) reaches
$100.0$ and KIVI-4 ($\approx31\%$ memory) reaches $99.5$---both within noise of the
uncompressed ceiling, confirming the quantization implementation is not silently
degrading accuracy on its own before any compression-specific comparison is drawn.
Second, Palu is run through a \texttt{llama3} RoPE shim so its published codec can be
evaluated on Llama-3.1 at all (Palu's original release targets an older RoPE
convention); the shim passes a sanity gate before any of its collapse numbers are
trusted---an uncompressed forward pass through the shimmed model reaches $99.5$ on
passage retrieval, against the true uncompressed ceiling of $100$, confirming the port
itself is correct and that Palu's subsequent collapse at aggressive budgets
(Appendix~\ref{app:palubudget}) is a genuine regime effect of the method, not an
artifact of an incorrect port. We also note, for completeness, a family of recent trained low-rank codecs.
MatryoshkaKV~\cite{matryoshkakv} and STAR-KV~\cite{starkv} adapt rank across layers,
heads, and blocks; DeltaKV~\cite{deltakv} encodes residuals against long-range
references; and DynaKV~\cite{dynakv} varies rank per token as VarRate does. All learn
their allocation through gradient training. VarRate is, to our knowledge, the
training-free point in this space, and we do not attempt a direct empirical comparison
against these methods here, since a training-free/trained contrast is a difference in
kind, not in accuracy at matched budget.

\subsection{The Published Low-Rank Codec: Full Budget Curve and Footprint Validation}
\label{app:palubudget}
The main paper's budget sweep against Palu (its Figure~\ref{fig:budget}) plots the same numbers
tabulated in full here (Table~\ref{tab:paludetail}) across four budgets and three
tasks. Palu was designed for the keep-$50$--$70\%$ regime; at keep-$20$--$30\%$ it
collapses to near-zero on passage and musique ($0.0$--$0.7$), and only becomes
viable once the budget loosens to keep-$50\%$ (passage $82.5$) and keep-$70\%$
(passage $97.0$)---exactly its own paper's operating range. VarRate strictly
dominates it at every budget tested, including keep-$70\%$, where Palu is closest to
competitive.

\begin{table*}[t]
\centering\small
\begin{adjustbox}{max width=\textwidth}
\begin{tabular}{lcccc}
\toprule
task (VarRate / Palu / flat) & keep-20\% & keep-30\% & keep-50\% & keep-70\% \\
\midrule
passage & 99.5 / 0.4 / 87.5 & 99.5 / 0.0 / 99.0 & 99.5 / 82.5 / 99.5 & 99.5 / 97.0 / 99.5 \\
musique & 32.0 / 0.7 / 30.9 & 33.2 / 0.5 / 31.7 & 33.6 / 21.0 / 34.0 & 34.3 / 25.3 / 34.3 \\
gov\_report & 30.4 / 3.4 / 30.1 & 32.7 / 3.2 / 32.6 & 34.5 / 21.0 / 34.4 & 34.5 / 33.1 / 34.5 \\
\bottomrule
\end{tabular}
\end{adjustbox}
\caption{Full Palu budget curve, Llama-3.1, $n{=}200$. Even our own uniform-rank flat
codec beats Palu at matched budget, isolating two separate gains: basis construction
(flat $\gg$ Palu) and salience allocation (VarRate $\gg$ flat).}
\label{tab:paludetail}
\end{table*}

That last observation is worth stating explicitly: our own flat, uniform-rank codec
already beats Palu at every matched budget in the table, before salience allocation
enters the picture at all. This separates two independent sources of VarRate's
advantage over Palu---one from how the shared low-rank basis itself is constructed
(flat's residual-to-anchor-mean codec versus Palu's direct low-rank projection), and
a second, larger one from allocating rank by salience rather than uniformly (the
comparison the main paper's own uniform-rank ablation isolates). Reporting Palu's
comparison at a memory-matched budget also required knowing its \emph{true} footprint
rather than its nominal one: computed directly from its released checkpoint's
per-layer \texttt{head\_wise\_ranks}, Palu's \texttt{ratio-0.2} configuration stores
$20.41\%$ of the cache (keys $7.4\%$, values $33.4\%$---its values are far less
compressible than its keys), marginally \emph{more} than VarRate's measured
$20.22\%$, confirming the keep-$20\%$ row above is memory-matched in Palu's favor if
anything. Palu cannot be run through the same measured-cost instrument as the other
baselines (it ships its own repository, environment, and fused kernel), but its
released accuracy pipeline reconstructs full K/V exactly as this paper's convention
does---\texttt{PaluLlamaForCausalLM} projects to a latent representation and then
immediately reconstructs it before caching, with the latent-only kernel reserved for
Palu's own separate latency microbenchmarks. The same accuracy-in-Mode-A,
efficiency-in-a-separate-kernel split holds for KIVI and other released low-rank/
quantization codecs we are aware of; it is the field's standard methodology, not a
limitation specific to this comparison. Finally, we attempted to reproduce Palu's own
latency claim on the GQA models used throughout this paper and could not: its fused
attention kernel assumes a multi-head attention (MHA) weight tensor
(\texttt{num\_heads}$\times$\texttt{rank}$\times$\texttt{head\_dim}) and raises when
given Llama-3.1-8B's GQA-shaped tensor (\texttt{num\_kv\_heads}$\times$\texttt{rank}
$\times$\texttt{head\_dim}) instead. Run in Palu's own MHA configuration, the kernel
does reproduce its claimed attention-module speedup ($1.28\times$ at 8K context,
$1.24\times$ at 16K, at a $50\%$ footprint)---so the failure is specific to GQA
models, consistent with Palu's paper benchmarking latency on an MHA model (Llama-2-7B)
and using a GQA model only for its accuracy numbers. Every model evaluated in this
paper (Llama-3.1, Qwen2.5, Mistral-7B) is GQA, which is why Palu's memory claim
carries over here but its kernel-level speedup claim does not.

\subsection{A Third Model at Full Budget Sweep (Mistral-7B)}
\label{app:mistral}
The main text reports that VarRate holds $73.6$ on Mistral-7B-Instruct-v0.2 retrieval
against flat's collapse to $29.1$, at a $15\%$ budget; Table~\ref{tab:mistral} gives
the full three-budget, seven-method comparison behind that summary. Mistral-v0.2 is
included for cross-validation and three-model-family breadth---it is Palu's own
LongBench evaluation model---but is a substantially weaker base model on these tasks
than Llama-3.1 or Qwen2.5 (uncompressed passage $75.6$, musique $16.9$, against
Llama's $100$ and $32$--$34$), so it is reported as a cross-check, not folded into the
two-model headline claim.

\begin{table*}[t]
\centering\small
\begin{adjustbox}{max width=\textwidth}
\begin{tabular}{lccc}
\toprule
method (KR15/20/25) & passage & musique & gov \\
\midrule
Full (ceiling) & 75.6 & 16.9 & 32.2 \\
SnapKV & 75.1 / 75.6 / 74.2 & 11.9 / 13.8 / 14.9 & 26.1 / 26.9 / 28.1 \\
PyramidKV & 74.3 / 75.7 / 76.7 & 15.8 / 16.9 / 16.1 & 25.4 / 26.7 / 27.3 \\
Ada-KV & 75.8 / 75.8 / 76.0 & 13.4 / 13.1 / 14.5 & 25.6 / 26.6 / 27.4 \\
flat (uniform rank) & 29.1 / 54.2 / 62.4 & 15.6 / 17.6 / 18.5 & 26.5 / 28.9 / 30.2 \\
\textbf{VarRate} & 73.6 / 73.3 / 74.0 & \textbf{17.7} / \textbf{17.8} / 17.1 & 27.7 / 29.3 / 30.7 \\
KIVI-2 / KIVI-4 & 73.1 / 76.6 & 16.8 / 17.3 & \textbf{32.4} / \textbf{32.2} \\
\bottomrule
\end{tabular}
\end{adjustbox}
\caption{Third-model cross-check, Mistral-7B-Instruct-v0.2, $n{=}200$. The same
pattern generalizes: flat collapses on retrieval, VarRate recovers to near-ceiling and
matches or beats full on multi-hop QA, and quantization again wins summarization.}
\label{tab:mistral}
\end{table*}

The pattern already established on Llama and Qwen generalizes cleanly to a third,
architecturally distinct and substantially weaker base model: the flat codec's
retrieval collapse (as low as $29.1$ at the most aggressive budget tested) is not a
Llama- or Qwen-specific artifact, VarRate's recovery to near-uncompressed accuracy on
retrieval ($73.3$--$74.0$, against a $75.6$ ceiling) holds, VarRate matches or exceeds
the uncompressed model on multi-hop QA (up to $17.8$ against a ceiling of $16.9$), and
quantization's summarization advantage over low-rank coding (KIVI $\approx32$ against
VarRate's $\approx30$) replicates as well.

\subsection{Calibration Robustness Across Three Independently Built Bases}
\label{app:calib}
The shared PCA basis each layer uses is built once, offline, from a handful of
unlabeled calibration contexts; if the codec's accuracy depended sensitively on which
contexts happened to be chosen, that would undermine its status as training-free and
broadly applicable. We rebuild the basis three separate times---the original
calibration set (v0), a set drawn from different examples (v1), and a set drawn from
different tasks entirely (v2)---and rerun both the flat codec and VarRate at KR20 on
all three (Table~\ref{tab:calib}).

\begin{table}[tb]
\centering\small
\begin{adjustbox}{max width=\columnwidth}
\begin{tabular}{lccc}
\toprule
method / task & v0 (original) & v1 (diff.\ examples) & v2 (diff.\ tasks) \\
\midrule
flat passage & 87.5 & 87.0 & 88.0 \\
flat musique & 30.9 & 32.5 & 31.8 \\
flat gov\_report & 30.1 & 30.2 & 30.4 \\
VarRate passage & 99.5 & 99.5 & 99.5 \\
VarRate musique & 32.0 & 33.8 & 33.5 \\
VarRate gov\_report & 30.4 & 30.6 & 30.0 \\
\bottomrule
\end{tabular}
\end{adjustbox}
\caption{Calibration robustness, Llama-3.1, KR20. Every spread across the three
independently built bases is $\leq\!1.8$ points, inside the confidence intervals
reported elsewhere in this paper.}
\label{tab:calib}
\end{table}

Every spread across the three bases, for both methods and all three tasks, is at
most $1.8$ points---inside the confidence intervals used throughout the rest of this
paper's significance testing (Appendix~\ref{app:sig}). The codec's accuracy is
therefore not tuned to, or dependent on, the specific calibration examples used to
build its basis; any small, unlabeled sample of representative contexts produces an
equivalent codec. The later negative-control, Qwen-signal, and floor experiments
(Apps.~\ref{app:controls}, \ref{app:signal-gen}, and \ref{app:floor}) use a
separately-built basis of this kind; VarRate's absolute scores there therefore sit
within this same spread of the main tables (e.g.\ Llama GovReport $30.0$ rather than
$30.4$), not exactly on them.

\subsection{A Data-Integrity Correction in the 16-Task Breadth Table}
\label{app:dataintegrity}
The main paper's 16-task LongBench breadth table (its Table~\ref{tab:main}) is assembled from
per-model, per-task result files by a small script rather than computed inline, and
an earlier version of that assembler had a real bug worth disclosing for
transparency. The original assembler globbed result files by pattern across all
models' output directories; because Qwen's and Mistral's runs reuse the same
\texttt{(method, task)} key names as Llama's, and the glob's file ordering was not
model-aware, later-sorted files from a different model silently overwrote the correct
Llama value for that key in $18$ of the table's $24$ reused cells---for example,
Mistral's \texttt{kivi\_k2v2/passage} score of $73.09$ overwrote Llama's true value of
$100.0$ for the same nominal key. The corrupted table happened to make VarRate look
like the top compressor and quantization look like it collapsed on retrieval, which
is precisely why it was caught rather than silently published: those two conclusions
contradicted the per-run JSON files each score was drawn from. The fix was to rebuild
the assembler from explicit, per-model-scoped source paths and to make it raise an
error on any duplicate key rather than silently overwrite, and to regenerate the
table from scratch. The individual per-run result files were never wrong---only the
now-corrected aggregation step was---and the earlier, narrower experiments in this
appendix (the three-task tables throughout) were assembled independently and were
unaffected.

\subsection{Query-Agnostic Budget Curve, and a Validated Expected-Attention Baseline}
\label{app:budgetcurve}
Figure~\ref{fig:pareto}'s keep-$20\%$ comparison could in principle be an artifact of
testing KVzip outside its own headline regime (keep-$25$--$33\%$). Sweeping keep-%
$\{10,20,25,50\}\%$ on Llama, memory-matched to within $\pm2\%$ at every point, VarRate
is at or above KVzip at keep-$20$, $25$, and $50\%$, including KVzip's own regime
(keep-$25\%$: $58.13$ vs.\ $56.83$). Only at keep-$10\%$ does the default stride fall
below KVzip, collapsing to $29.92$ (App.~\ref{app:stride} shows this is a stride artifact, not a
property of low-rank coding: widening the stride to $64$ recovers $48.88$, above
KVzip's $42.12$). Expected Attention is separately validated against its own paper's
claimed regime---we reproduce its reported near-parity at keep-$50\%$ query-aware
($-1.2$ mean)---so its $-21.7$ collapse under query-agnostic reuse is a genuine
mechanistic finding (its future-query estimate is built from a prefill that normally
contains the question) and not a misconfiguration on our part. We additionally check
Expected Attention against the exact regime its own paper claims: at keep-$50\%$,
query-aware, it reports near-parity with the uncompressed model, and we reproduce that
directly (passage $100.0\to98.0$, musique $32.76\to32.56$, qasper $47.2\to45.94$, a
$-1.2$ mean drop). Its published evaluation also sweeps removal ratios of
$0/10/25/50/75/90\%$, which brackets our $80\%$-removed ($\kappa{=}0.20$) operating
point, so the $-21.7$ collapse we observe under query-agnostic reuse is squarely
inside its own tested range---not an extrapolation to an untested regime where a
method could reasonably be expected to fail. Taken together, this rules out the two
easiest objections to Figure~\ref{fig:pareto} and Table~\ref{tab:agnostic}: that
KVzip was tested outside its comparative advantage, and that Expected Attention's
weakness is an artifact of our harness rather than a property of the method itself.

\begin{table*}[t]
\centering\small
\begin{adjustbox}{max width=\textwidth}
\begin{tabular}{lccc}
\toprule
keep \% & VarRate & KVzip & Expected Attention \\
\midrule
50\% & \textbf{60.81} & 58.66 & 55.55 \\
25\% (KVzip's own regime) & \textbf{58.13} & 56.83 & 37.58 \\
20\% & \textbf{56.21} & 55.34 & 32.83 \\
10\% & 29.92 (stride 16) / \textbf{48.88} (stride 64) & 42.12 & 23.23 \\
\bottomrule
\end{tabular}
\end{adjustbox}
\caption{Query-agnostic budget curve, Llama-3.1, 3-task mean. VarRate $\geq$ KVzip at
keep-$20/25/50\%$, including KVzip's own 25\% regime; at keep-10\% the default stride
(s16) falls below, a stride artifact (App.~\ref{app:stride}) that widening the stride
to s64 recovers.}
\label{tab:budgetcurve}
\end{table*}

\subsection{Where the Prefill Overhead Goes, and How Far It Compresses}
\label{app:profile}
Profiling \texttt{compress()} stage-by-stage (CUDA events, summed over 32 layers,
cross-checked against wall-clock to $1.8\%$) attributes the $+44.0\%$ overhead almost
entirely to implementation, not arithmetic: reference-token search ($32.5\%$ of
compress time, $0.84\%$ of its FLOPs), the PCA projection ($21.8\%$ time,
$1.67\%$ FLOPs), reconstruction, salience scoring, and water-filling together account
for the rest, with a materialized fp32 distance matrix, thousands of small kernel
launches from water-filling, and a per-layer CPU$\to$GPU basis re-upload as the actual
costs. A separate, bit-identity-controlled implementation
(\texttt{varrate\_fast\_press.py}) isolates this on its own timing run---its reference
reads $+41.7\%$, against Table~\ref{tab:cost}'s $+44.0\%$: caching the basis on GPU and
using an analytic reference count reproduces the reference exactly ($99.50/31.97/30.45$,
bit-identical) while cutting it to $+40.3\%$; adding bf16 compute
on top cuts it further to $+25.1\%$, with accuracy unchanged (max $|\Delta|=0.07$,
inside noise). The remaining floor is $+2.6\%$ of prefill in pure arithmetic---the
reference implementation
in the main tables is not close to that floor, and a fused kernel is the natural next
step. The profile itself is cross-validated two ways: the sum of the six stage
timings ($732.3$ms) agrees with an independently measured wall-clock delta between
compressed and uncompressed runs ($719.2$ms) to within $1.8\%$, so the breakdown is
not an artifact of how the instrumentation itself perturbs timing. We also record two
optimizations that were tried and explicitly rejected, because both are easy mistakes
that silently corrupt accuracy rather than merely underperforming: computing the
anchor-distance matrix in bf16 (via the $\lVert a\rVert^2+\lVert b\rVert^2-2ab$
identity) catastrophically cancels, producing a $15.5\%$ mean relative error in the
reference means and is kept in fp32 for that reason; and an early bf16 water-filling
variant fed the solver fp32 salience scores where the shipped reference always uses
bf16 scores (matching SnapKV's native precision), which silently changed the resulting
rank allocation and broke bit-identity by $0.10$ points on musique before being
caught and fixed. Both are recorded so a reimplementation does not rediscover them
the expensive way.

\begin{table}[tb]
\centering\small
\begin{adjustbox}{max width=\columnwidth}
\begin{tabular}{lccc}
\toprule
stage & \% of compress() & \% of prefill & \% of FLOPs \\
\midrule
reference search (cdist+topk) & 32.5\% & 14.9\% & 0.84\% \\
PCA projection & 21.8\% & 10.0\% & 1.67\% \\
reconstruction & 19.9\% & 9.1\% & --- \\
salience scoring & 10.7\% & 4.9\% & 0.10\% \\
water-filling & 6.4\% & 2.9\% & $\sim$0 \\
un-/re-rotate & 8.6\% & 3.9\% & --- \\
\bottomrule
\end{tabular}
\end{adjustbox}
\caption{Stage-by-stage profile of \texttt{compress()} (32 layers, 16K ctx, A100).}
\label{tab:profile}
\end{table}

\begin{table}[tb]
\centering\small
\begin{adjustbox}{max width=\columnwidth}
\begin{tabular}{lcccc}
\toprule
variant & prefill $\Delta$ & passage & musique & gov \\
\midrule
reference & $+41.7\%$ & 99.50 & 31.97 & 30.45 \\
fast\_fp32 (GPU-cached basis + analytic $n_{\mathrm{ref}}$) & $+40.3\%$ & 99.50 & 31.97 & 30.45 \\
fast\_bf16 (+ bf16 compute) & $\mathbf{+25.1\%}$ & 99.50 & 31.90 & 30.51 \\
\bottomrule
\end{tabular}
\end{adjustbox}
\caption{Optimized implementations, bit-identity controlled. \texttt{fast\_fp32} is
exactly reference; \texttt{fast\_bf16} cuts overhead $40\%$ with accuracy unchanged.}
\label{tab:fastimpl}
\end{table}

\subsection{The Stride Ablation}
\label{app:stride}

\begin{figure*}[!t]
\centering
\includegraphics[width=\textwidth]{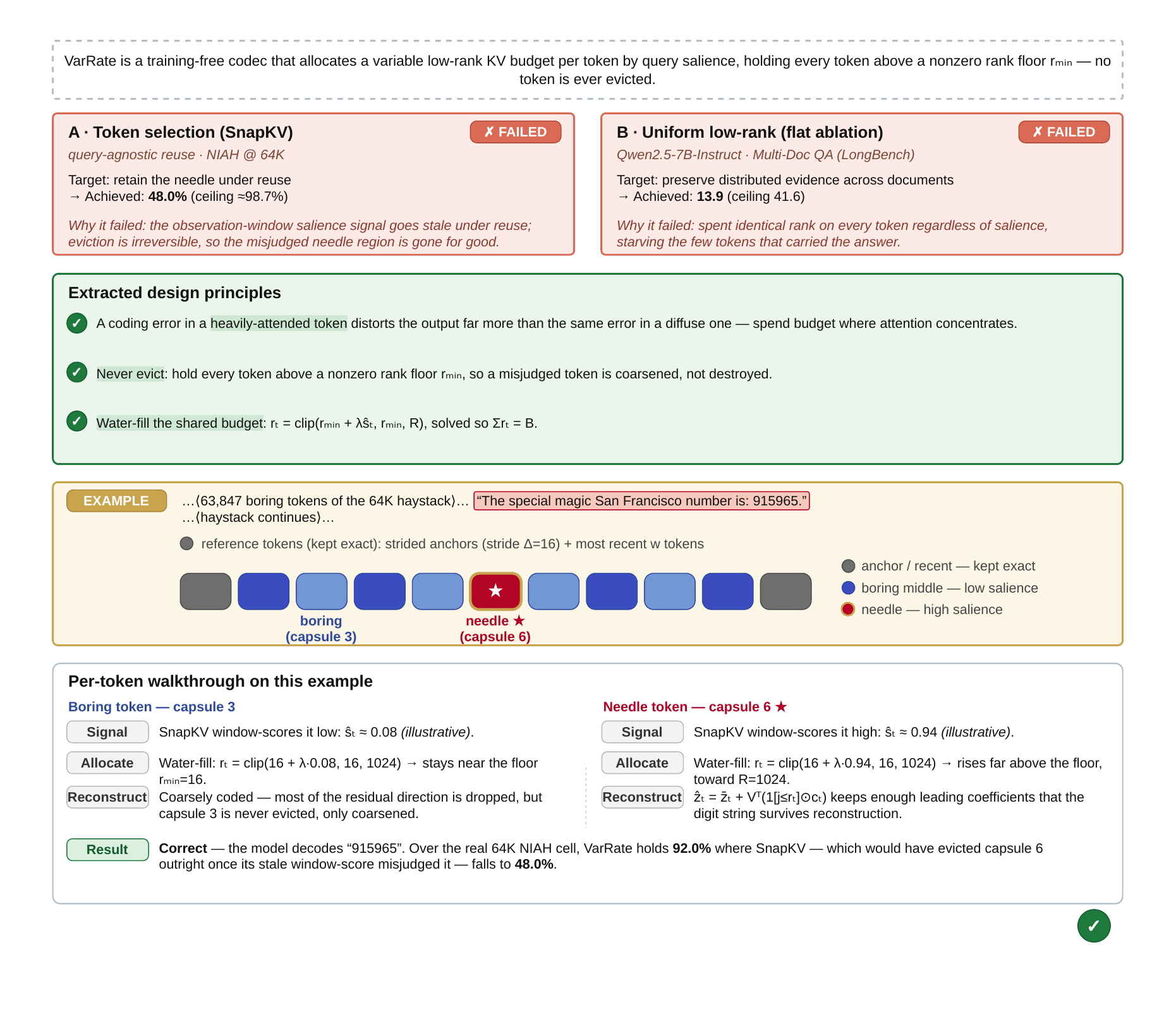}
\caption{\textbf{Worked example of the per-token allocation on a $64$K needle-in-a-haystack
instance.} Each token's salience $\hat{s}_t$ (its SnapKV observation-window score) sets a per-token
rank $r_t$ by water-filling the shared budget: a low-salience \emph{boring} token is coarsened to the
floor $r_{\min}$, whereas the high-salience \emph{needle} token is raised toward the cap $R$, so that
its key and value survive reconstruction. This is the mechanism behind VarRate's $92.0\%$ at the
$64$K cell, where SnapKV---having irreversibly evicted the needle once its stale window score
misjudged it---falls to $48.0\%$ (Table~\ref{tab:niah}). The salience values shown are illustrative.}
\label{fig:walkthrough}
\end{figure*}

VarRate stores strided reference tokens at full width (the exact-kept anchors in
Figure~\ref{fig:walkthrough}, which walks the full per-token scheme through one needle); at the
default stride $16$ they
are $6.25\%$ of the cache, so at an aggressive keep-$10\%$ budget they alone consume
$\sim\!66\%$ of it. Widening the stride re-splits the (stride-invariant) footprint
toward coefficients: at keep-$10\%$, query-agnostic, stride $16\to64$ swings
$+18.96$ points ($29.92\to48.88$), turning a loss to KVzip into a win. This does
\emph{not} generalize to the full 16-task suite, however: at the main keep-$20\%$
operating point, stride $64$ costs $-0.71$ overall (AVG-16 $44.98$ vs.\ $45.69$),
concentrated in few-shot ($-3.1$) and code ($-2.69$)---structured contexts whose
exemplars and cross-file references must survive verbatim, which quartering the exact
token budget destroys. Multi-hop QA gains slightly ($+0.8$) because the freed budget
raises average rank. The trade is explicit: stride governs structural preservation
versus reconstruction fidelity, and the optimum shifts with the budget and the task
type. We keep stride$=16$ as the committed default and report the wider setting only
as an ablation, never tuned on the eval tasks themselves. The accept/reject
methodology matters here as much as the numbers: the query-agnostic 3-task signal at
keep-$20\%$ ($+1.13$, inside its own confidence interval) would, on its own, have
looked like a plausible case for switching the default---and would have led us to
rerun the full 16-task suite with the \emph{worse} setting had that decision been made
after seeing the smaller, favorable number. The rule for when a hyperparameter
ablation earns a full 256-example, 16-task rerun was fixed before that rerun's result
was known, precisely to prevent this kind of favorable-looking-in-isolation number
from silently becoming the shipped default. Stride is reported here as exactly what
it is: a real trade-off between how many tokens are stored exactly and how much rank
budget survives for everyone else, with no universally correct setting, rather than a
knob tuned post hoc on the evaluation tasks.

\begin{table}[tb]
\centering\small
\begin{adjustbox}{max width=\columnwidth}
\begin{tabular}{lcc}
\toprule
keep \% & stride 16 (default) & stride 64 \\
\midrule
10\% (query-agnostic, 3-task mean) & 29.92 & \textbf{48.88} \\
20\% (query-agnostic, control) & 56.21 & 57.34 \\
20\% (AVG-16, full suite) & \textbf{45.69} & 44.98 ($-0.71$) \\
\bottomrule
\end{tabular}
\end{adjustbox}
\caption{Stride trades structural preservation against reconstruction fidelity:
wider stride wins at an aggressive 10\% budget but loses $0.71$ on the full 16-task
suite (few-shot $-3.1$, code $-2.69$), so stride$=16$ stays the default.}
\label{tab:stride}
\end{table}

\subsection{Statistical Significance, Summarized}
\label{app:sig}
Every comparison in the main text is a paired bootstrap (20{,}000 resamples, 95\% CI)
over the 16 LongBench tasks, backed by an exact sign test; Table~\ref{tab:sig}
summarizes the verdicts referenced throughout. The pattern that survives testing is:
VarRate never loses significantly to a matched-memory baseline on either model, wins
significantly against selection only on Qwen, ties quantization on both models, and
beats its own uniform-rank ablation significantly on both---the strongest and most
robust result in the paper. We treat a comparison as supported only when the paired
bootstrap CI excludes zero; a nominal mean difference in VarRate's favor that does not
clear that bar (e.g.\ Ada-KV on Llama, $+0.36$, or KIVI-2 on either model) is reported
as a tie, not as a win, throughout the paper, and the reverse discipline applies
equally---KIVI-4's nominal $-0.98$ on Qwen is a real, CI-excluding loss, but at
$1.55\times$ the memory of every other method in the comparison, so it is reported as
significant but not memory-matched rather than folded into an unqualified ``VarRate
loses to KIVI'' claim. This two-sided discipline is what lets the single strongest
number in the table---VarRate beating its own uniform-rank ablation by $+2.22$ (Llama)
and $+17.50$ (Qwen), both significant at $p\leq0.021$ and winning $13/16$ and $15/16$
tasks respectively---stand as the paper's central claim rather than as one cherry
among several nominal wins.

\begin{table*}[t]
\centering\small
\begin{adjustbox}{max width=\textwidth}
\begin{tabular}{lcccc}
\toprule
baseline & Llama $\Delta$ & Llama 95\% CI & Qwen $\Delta$ & Qwen 95\% CI \\
\midrule
PyramidKV & $-0.33$ & [$-2.03$, $+1.16$] tie & $\mathbf{+3.10}$ & [$+1.36$, $+5.75$]$^{*}$ \\
SnapKV & $-0.10$ & [$-1.18$, $+0.93$] tie & $\mathbf{+1.75}$ & [$+0.49$, $+3.59$]$^{*}$ \\
Ada-KV & $+0.36$ & [$-0.64$, $+1.37$] tie & $+1.52$ & [$+0.51$, $+2.81$]$^{*}$ \\
KIVI-2 & $+0.86$ & [$-0.73$, $+2.76$] tie & $+1.18$ & [$-0.30$, $+3.11$] tie \\
\textbf{flat (ours)} & $\mathbf{+2.22}$ & [$+0.72$, $+4.05$]$^{*}$ & $\mathbf{+17.50}$ & [$+10.77$, $+24.69$]$^{*}$ \\
\bottomrule
\end{tabular}
\end{adjustbox}
\caption{Paired VarRate$-$baseline deltas over 16 tasks, 20{,}000-sample bootstrap.
$^{*}$CI excludes zero.}
\label{tab:sig}
\end{table*}

\subsection{Real Wall-Clock Latency}
\label{app:latency}
Table~\ref{tab:cost}'s prefill overhead is an accounted, not measured, estimate for
this section's purpose; we additionally time full end-to-end \texttt{pipe()} calls
(n=18, 2 warmup runs discarded) on one prefill-dominated task (short-answer retrieval)
and one decode-dominated task (long-generation summarization). On retrieval, VarRate's
overhead is $+41.8\%$ end-to-end, at the top of the accounted range, because
\texttt{compress()} is a fixed one-time prefill cost and prefill dominates a short
generation. On summarization, the identical fixed cost dilutes to $+2.6\%$, within
run-to-run noise ($\pm4.36$s)---indistinguishable from uncompressed. SnapKV and KIVI
stay near-zero overhead on both tasks, as expected from their respective mechanisms:
SnapKV's scoring is a single additional attention read over the observation window,
and KIVI's quantization is a fixed-cost per-tensor quantize step, neither of which
scales with the codec-style projection and reconstruction work VarRate performs. The
practical reading: VarRate's compute tax is concentrated in short-generation,
retrieval-style workloads and disappears for long-generation workloads, the inverse of
where its accuracy advantage over selection is smallest (Appendix~\ref{app:selfam}
shows the summarization win, which is exactly the long-generation regime where the
latency tax is noise). A deployment sensitive to prefill latency on short-answer
workloads should weigh this tax against the accuracy and robustness gains reported
elsewhere in this appendix; a deployment dominated by long-generation summarization or
multi-turn dialogue pays close to nothing for the same benefits.

\begin{table*}[t]
\centering\small
\begin{adjustbox}{max width=\textwidth}
\begin{tabular}{llcc}
\toprule
task (mean ctx) & config & total (s) & $\Delta$ vs.\ full \\
\midrule
passage\_retrieval\_en (12.6K tok) & Full & 1.256 & --- \\
& SnapKV @KR20 & 1.280 & $+1.9\%$ \\
& \textbf{VarRate @KR20} & \textbf{1.781} & $\mathbf{+41.8\%}$ \\
& KIVI-2 & 1.340 & $+6.7\%$ \\
gov\_report (14.7K tok) & Full & 13.59 & --- \\
& SnapKV @KR20 & 13.42 & $-1.3\%$ (noise) \\
& \textbf{VarRate @KR20} & \textbf{13.95} & $\mathbf{+2.6\%}$ (noise) \\
& KIVI-2 & 14.01 & $+3.1\%$ \\
\bottomrule
\end{tabular}
\end{adjustbox}
\caption{Real end-to-end wall-clock, Llama-3.1-8B, one A100 ($n{=}18$). VarRate's
fixed \texttt{compress()} cost dominates on prefill-heavy retrieval but is noise on
decode-heavy summarization.}
\label{tab:latency}
\end{table*}

\subsection{A Second Long-Context Suite (RULER), Limited Scope}
\label{app:ruler}
As a task-diversity cross-check beyond the three tasks used throughout, we also ran
RULER's~\cite{hsieh2024ruler} synthetic suite (niah / QA / multi-hop / aggregation) at ctx$=16$K, $n=40$
(Table~\ref{tab:ruler}). The gap between VarRate and SnapKV ($+0.8$) is within noise at
this $n$; both compressors give up a real $\sim\!8$ points against uncompressed on
RULER's harder mixed suite. The available HF mirror only hosts 4K/8K/16K subsets, so
this is a diversity check, not the long-context scaling result---that claim is made
instead by the controlled needle-in-a-haystack (NIAH) sweep to $128$K (App.~\ref{app:niah}). RULER's four
task families are harder and more synthetically adversarial than the three real-text
tasks (passage retrieval, multi-hop QA, summarization) used throughout the rest of
this paper, so a consistent VarRate/SnapKV ordering here is evidence that the main
results are not an artifact of always reusing the same three tasks. We had originally
scoped this as a $16$K--$128$K scaling stress test, matching the NIAH sweep in
App.~\ref{app:niah}; that plan was dropped mid-run once we confirmed, via the hosted
repository's own file listing rather than a failed download, that the HF mirror used
here (\texttt{simonjegou/ruler}) simply does not host $32768/65536/131072$-token
subsets at all. We report the $16$K result that was already run rather than discard
it, but flag explicitly that it does not substitute for a scaling claim.

\begin{table}[tb]
\centering\small
\begin{adjustbox}{max width=\columnwidth}
\begin{tabular}{lc}
\toprule
config & RULER score (ctx=16K) \\
\midrule
Full & $90.0\pm9.30$ \\
SnapKV @KR20 & $81.7\pm11.77$ \\
\textbf{VarRate @KR20} & $82.5\pm11.78$ \\
\bottomrule
\end{tabular}
\end{adjustbox}
\caption{RULER cross-check, Llama-3.1-8B, $n=40$. VarRate/SnapKV gap is within noise.}
\label{tab:ruler}
\end{table}

\subsection{A Powered Re-Gate of the KVzip Comparison}
\label{app:kvzip-pex}
The main text's KVzip verdicts rest on a $3$-task mean, whose sign test cannot reach
significance in principle (floor $p=0.25$ at $n=3$). We additionally pair VarRate
against KVzip \emph{per example}, pooled over the three tasks ($\sim\!600$ paired
examples/cell, verified by a gold-answer fingerprint), summarized in
Table~\ref{tab:kvzippex}. This confirms, with real statistics rather than an eyeballed
CI band, that VarRate carries no accuracy advantage over KVzip and is significantly
behind only in one regime. In the same per-example analysis, VarRate beats every
selection/flat baseline in the query-agnostic regime with significance (vs.\ SnapKV
$+10.2$/$+9.9$, vs.\ Ada-KV $+8.2$/$+11.6$, vs.\ flat $+14.5$/$+29.3$, Llama/Qwen). The
per-example pairing is necessary rather than cosmetic: a $3$-task mean has an exact
two-sided sign test whose smallest achievable $p$-value at $n{=}3$ is $0.25$, so no
amount of margin on a 3-task table can ever cross a conventional significance
threshold, no matter how large or consistent the underlying effect is. Pairing at the
level of individual examples restores real statistical power without changing what is
being measured, as long as the pairing itself is trustworthy; we verify that
VarRate's and KVzip's per-example scores are aligned to the same underlying question
by matching on a fingerprint of the gold answer, rather than assuming the two
evaluation runs iterated examples in the same order. This is what allows the main
text's ``no accuracy advantage over KVzip, advantage is cost'' claim to rest on a real
confidence interval rather than an eyeballed $\pm2.5$ band on three numbers.

\begin{table*}[t]
\centering\small
\begin{adjustbox}{max width=\textwidth}
\begin{tabular}{lccc}
\toprule
cell & VarRate$-$KVzip & 95\% CI ($n{\approx}600$/cell) & verdict \\
\midrule
Llama, query-aware & $-0.55$ & [$-2.00$, $+0.91$] & tie \\
Llama, query-agnostic & $+0.95$ & [$-1.48$, $+3.42$] & tie \\
Qwen, query-aware & $-0.04$ & [$-1.67$, $+1.62$] & tie \\
Qwen, query-agnostic & $-5.06$ & [$-7.57$, $-2.58$] & KVzip$^{*}$ \\
\midrule
pooled (all 4 cells, $n{=}2400$) & $-1.18$ & [$-2.20$, $-0.16$] & KVzip$^{*}$ \\
\bottomrule
\end{tabular}
\end{adjustbox}
\caption{Per-example paired bootstrap, VarRate vs.\ KVzip, pooled over 3 tasks.
$^{*}$CI excludes zero. 3 of 4 cells are powered ties; KVzip leads only in Qwen
query-agnostic reuse.}
\label{tab:kvzippex}
\end{table*}

\subsection{Is the Signal Content, or Just Shape?}
\label{app:controls}
Table~\ref{tab:ablation}'s random signal changes the rank \emph{marginal} along with the
alignment, leaving open whether a graded, recency-like profile would do just as well
without reading content at all. Two further controls isolate this: \emph{shuffle}
permutes the real salience scores across coded tokens (same marginal, alignment
destroyed), and \emph{position} keeps only the recency-and-attention-sink shape that
Appendix~\ref{app:floor}'s rank logs reveal, with no content signal at all. On Qwen,
where low-rank coding is least forgiving, position collapses on multi-hop QA ($7.7$
vs.\ mass's $28.7$) and shuffle drops \emph{below} flat on both retrieval ($64.7$ vs.\
$75.0$) and multi-hop QA ($7.5$ vs.\ $8.5$)---a non-uniform allocation aimed at the
wrong tokens is worse than a uniform one. Content alignment, not gradedness alone, is
what the signal contributes; recency-with-extra-steps is refuted as an alternative
explanation.

\begin{figure*}[t]
\centering
\includegraphics[width=\textwidth]{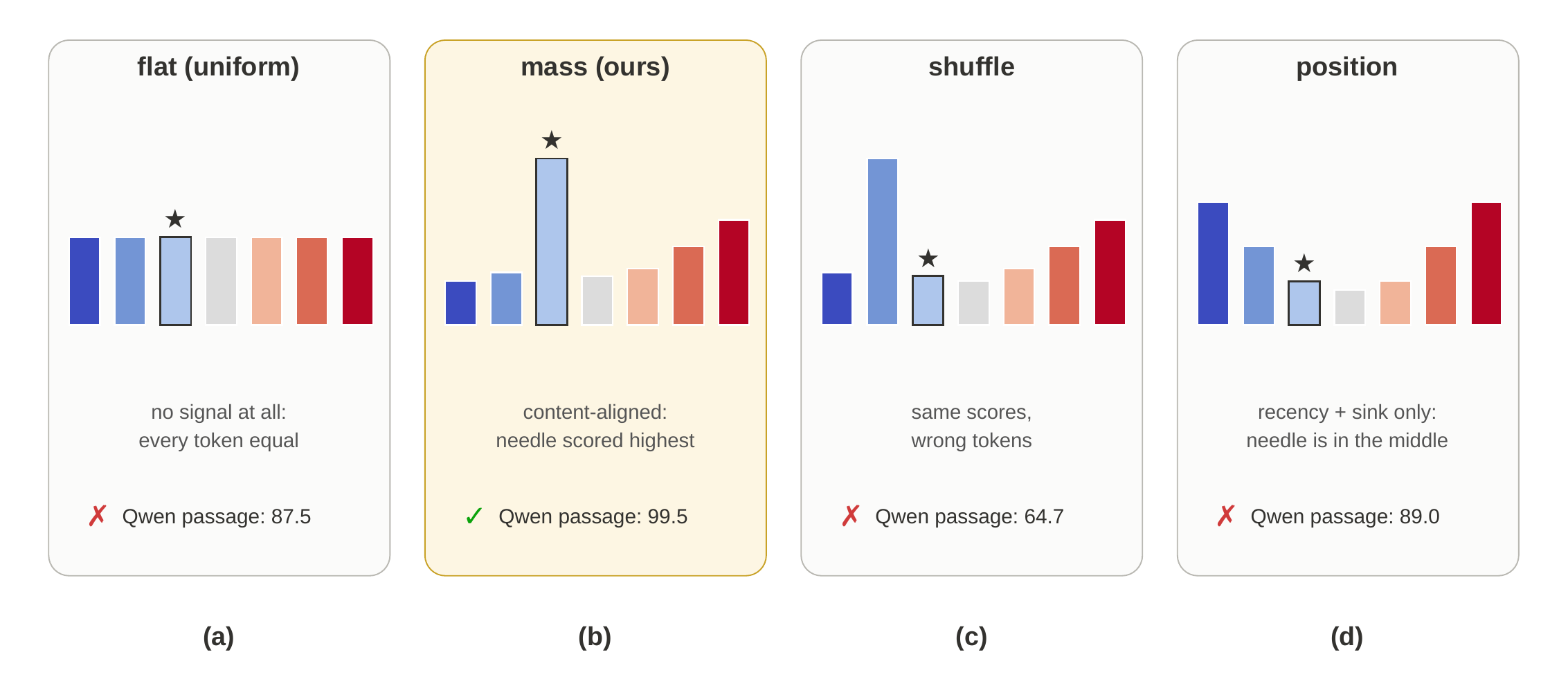}
\caption{\textbf{Why the signal has to be content-aligned, not just graded.} One
token (\textcolor{black}{$\bigstar$}, outlined) plays a query-critical ``needle''
placed in the document's boring middle, away from both the attention-sink start and
the query-proximate end. \textbf{(a)} Flat coding does not look at the needle at all.
\textbf{(b)} Mass (ours) scores it highest, because it is the only condition that
reads content. \textbf{(c)} Shuffle keeps the identical multiset of scores from (b)
but reassigns them across tokens, so the needle inherits an arbitrary score instead of
its own. \textbf{(d)} Position reproduces the recency/attention-sink \emph{shape} from
Figure~\ref{fig:decile} with no content signal, so a needle in the middle is scored
low purely by where it sits. Only the content-aligned condition (b) recovers the
needle; a graded-but-content-free score, however shaped, does not substitute for it.}
\label{fig:controls}
\end{figure*}

Figure~\ref{fig:controls} makes the mechanism concrete with a single traced token: a
schematic ``needle'' sitting in the document's boring middle (Figure~\ref{fig:decile}'s
low-rank zone), where recency and attention-sink effects are weakest and content
alignment is the only thing that can still find it. \emph{mass} is the only condition
that scores this token highest and recovers it; \emph{shuffle} keeps the exact same
set of scores but attaches them to the wrong tokens, so the needle gets whatever
another token's score happened to be; \emph{position} never looks at content at all,
so a needle sitting away from both edges is scored low regardless of how critical it
is. The bar heights are illustrative of the mechanism, not a plot of the ablation
itself---Table~\ref{tab:controls} below gives the real numbers behind each panel's
accuracy annotation.

\begin{table*}[t]
\centering\small
\begin{adjustbox}{max width=\textwidth}
\begin{tabular}{lccc}
\toprule
config & Llama (passage / gov / musique) & Qwen (passage / gov / musique) \\
\midrule
flat (uniform) & 87.5 / 30.1 / 30.7 & 75.0 / 29.6 / 8.5 \\
\textbf{VarRate (mass)} & \textbf{99.5 / 30.0 / 31.8} & \textbf{99.5 / 31.3 / 28.7} \\
shuffle (marginal held, alignment destroyed) & 93.5 / 28.7 / 30.4 & 64.7 / 27.4 / 7.5 \\
position (recency+sink shape, no content) & 86.5 / 29.7 / 30.5 & 89.0 / 26.5 / 7.7 \\
\bottomrule
\end{tabular}
\end{adjustbox}
\caption{Negative controls, KR20, $n{=}200$. \emph{shuffle} holds the rank marginal but
destroys alignment; \emph{position} is recency+sink shape with no content. Both
underperform mass, and on Qwen both underperform flat.}
\label{tab:controls}
\end{table*}

\subsection{Does the Signal Generalize Past One Heuristic and One Model?}
\label{app:signal-gen}
Replacing SnapKV attention with a structurally different second-order signal---attention
squared, weighted by value-distinctiveness through $W_O$, adapted and reindexed from a
closed-form Fisher-sensitivity metric developed for offline basis whitening rather than
online per-token rank allocation---ties mass on Llama, and on Qwen (where the signal
axis bites hardest) it ties again. This second-order signal is still computed from a
single forward pass with no gradients, so the comparison is between two training-free
heuristics, not between a heuristic and a trained alternative. The result is not an
artifact of one particular salience heuristic: a signal built from a completely
different mathematical quantity (second-order attention sensitivity rather than
first-order attention mass) lands within the confidence interval of the default on
every task, on both models, which is the strongest form of evidence available that
`mass' was not a lucky or overfit choice among many salience heuristics we could have
picked. The allocation axis replicates too: on Qwen, water-filling beats a fixed binary
split on every task that is not pure retrieval, so Llama's one point in fixed-split's
favor does not carry over (Table~\ref{tab:qwensig})---if anything, Qwen is the sharper
test of the allocation choice, because its four-KV-head cache is far less forgiving of
a wrong allocation shape than Llama's, as the main uniform-rank ablation already
shows, making any advantage for fixed-split easier to see if it existed.

\begin{table}[tb]
\centering\small
\begin{adjustbox}{max width=\columnwidth}
\begin{tabular}{lccc}
\toprule
config (Qwen, KR20) & passage & gov\_report & musique \\
\midrule
flat (uniform) & 75.0 & 29.6 & 8.5 \\
random signal & 71.6 & 29.1 & 6.5 \\
mass + fixed-split & 99.8 & 29.4 & 26.5 \\
fisher + waterfill & 100.0 & \textbf{30.8} & 27.9 \\
\textbf{mass + waterfill (ours)} & 99.5 & \textbf{31.3} & \textbf{28.7} \\
\bottomrule
\end{tabular}
\end{adjustbox}
\caption{Signal and allocation axes replicated on Qwen2.5-7B. Fisher ties mass; on
Qwen, unlike Llama, water-filling beats fixed-split on every non-retrieval task.}
\label{tab:qwensig}
\end{table}

\subsection{Is the Rank Floor a Lever?}
\label{app:floor}
\begin{figure*}[t]
\centering
\includegraphics[width=\textwidth]{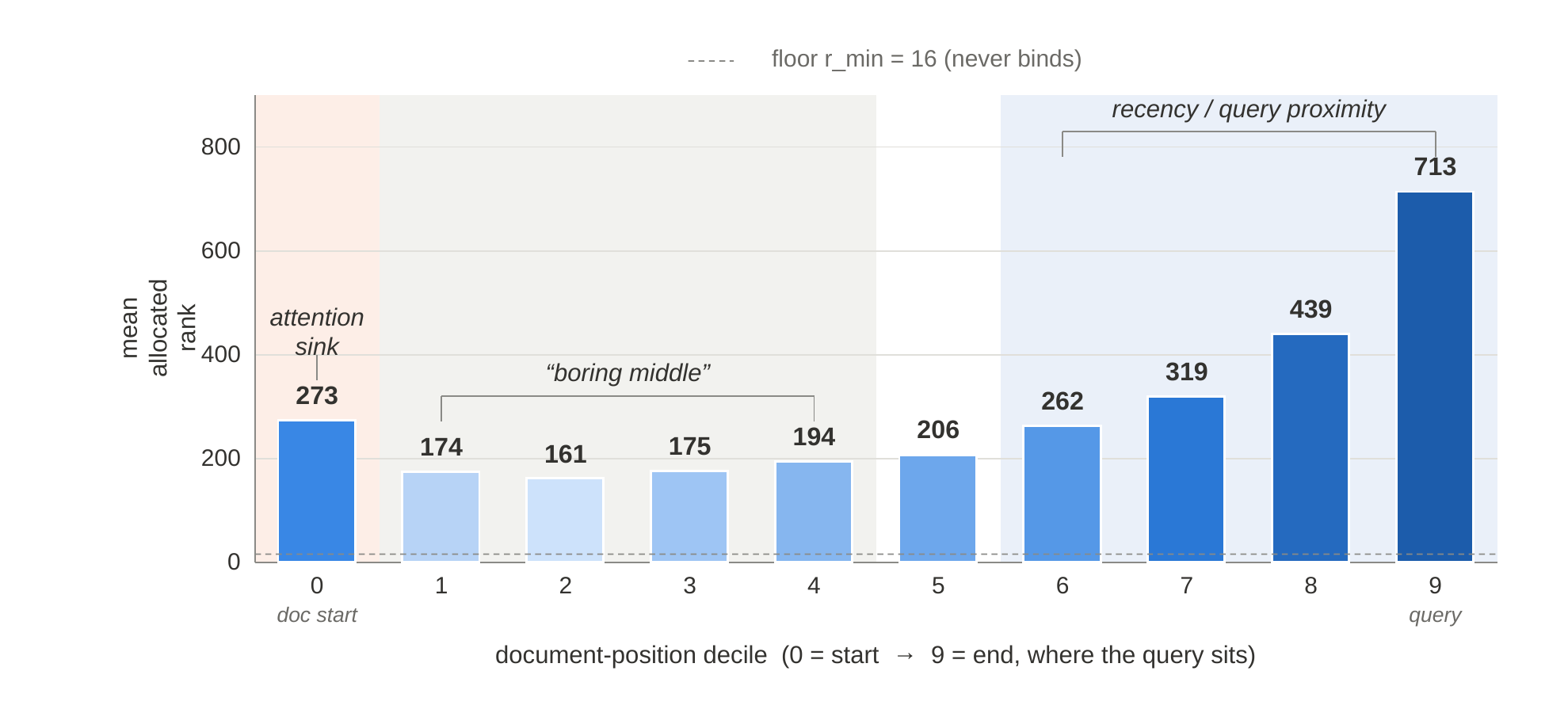}
\caption{\textbf{Where does the rank budget actually go?} Mean allocated rank by
document-position decile, pooled over $300$ sampled (layer, example) traces on
gov\_report at KR20 (0 = document start, 9 = the end, where the query sits after the
query-aware transform appends it). Three regimes, annotated directly on the data:
an \emph{attention-sink} bump at the very start (a well-documented effect where the
first tokens draw disproportionate attention regardless of content); a dip through
the ``boring middle,'' where content alignment (Appendix~\ref{app:controls}) is doing
most of the remaining work; and a sharp, monotonic climb toward the query at the end,
where recency and query proximity dominate. The dashed floor line is the same
$r_{\min}=16$ from Table~\ref{tab:rmin}---every decile's \emph{mean} sits $10$--$45\times$
above it, visual confirmation that the floor is slack everywhere, not just on average.}
\label{fig:decile}
\end{figure*}

Setting $r_{\min}=0$ changes accuracy by less than a point on every task
(Table~\ref{tab:rmin}), which could mean water-filling rediscovers hard eviction once
permitted to. It does not: per-token rank logs over $3$--$4.3$M coded tokens show
$0.0\%$ of tokens at the floor under either setting, at both the main $20\%$ budget
and, extending the check, at $10\%$ and $15\%$. The floor never binds; VarRate is not
secretly falling back to selection. The same logs surface a mechanistic finding
instead (Figure~\ref{fig:decile}): allocated rank is elevated at the very start of the
document (an attention-sink effect), dips through the middle, and climbs sharply
toward the query at the end---the signal is dominated by recency and query proximity,
with content alignment (Appendix~\ref{app:controls}) supplying the rest. This was
originally motivated by a stronger hypothesis: that water-filling would naturally
\emph{rediscover} hard eviction for the least-salient tokens once the floor no longer
forbids it, which would make VarRate a strict generalization subsuming both flat
low-rank coding and token selection as special cases. The rank logs refute that
stronger claim directly---the floor sits at $0.0\%$ occupancy whether or not it is
even present, so there is no mass of low-salience tokens straining against
$r_{\min}{=}16$ that gets released to zero once the constraint is lifted. The correct
description is more modest but still useful: at this operating point, water-filling's
natural allocation already keeps every coded token comfortably above the floor, so the
floor functions as an unused safety margin rather than as an active mechanism, and the
``strict generalization'' framing should be dropped in favor of a claim about accuracy
parity under floor removal. A related diagnostic is worth flagging honestly: a pooled
Pearson correlation between raw, un-normalized salience scores and final allocated
rank across all sampled (layer, example) traces comes out surprisingly low
($\sim\!0.05$--$0.06$). This is very likely a pooling artifact rather than evidence
that salience weakly predicts rank---raw attention-mass scale varies substantially
across different forward passes, and water-filling only enforces a monotonic
salience-rank relationship \emph{within} a single pass, after per-pass min--max
normalization, not across pooled raw scores drawn from different passes. We do not read it as contradicting the main text's oracle comparison and signal
ablation, both of which show that salience does predict rank well; it simply shows
that this particular pooled statistic is the wrong instrument to confirm it, and a
within-pass rank correlation would be the correct follow-up if this needs to be
demonstrated more directly.

\begin{table}[tb]
\centering\small
\begin{adjustbox}{max width=\columnwidth}
\begin{tabular}{lccc}
\toprule
task (Llama, KR20) & $r_{\min}{=}16$ (default) & $r_{\min}{=}0$ & $\Delta$ \\
\midrule
passage\_retrieval\_en & 99.5 & 99.5 & 0.0 \\
gov\_report & 30.04 & 30.64 & $+0.6$ \\
musique & 31.76 & 31.29 & $-0.47$ \\
\bottomrule
\end{tabular}
\end{adjustbox}
\caption{Removing the rank floor costs nothing, but per-token logs show $0.0\%$ of
coded tokens at the floor under either setting---the floor is slack, never binding.}
\label{tab:rmin}
\end{table}

\subsection{Does the Advantage Grow at Aggressive Budgets?}
\label{app:aggbudget}
The main results operate at keep-$15$--$25\%$; a natural conjecture is that VarRate's
margin over selection grows as the budget tightens further. Sweeping to keep-$8$--$12\%$
(Table~\ref{tab:aggbudget}) falsifies this. At keep-$8\%$, SnapKV matches or beats
VarRate on both models, and on Qwen VarRate collapses to flat's level: below a
viability threshold, exact selection keeps the query-relevant tokens on any budget,
while a shared low-rank basis cannot. VarRate recovers parity with selection by
keep-$10\%$ (Llama) and keep-$12\%$ (Qwen), and still beats the flat codec throughout.
This is the mirror image of the query-agnostic result in the main text: which method
wins is set by the \emph{regime}---query-aware versus reused---not simply by how
aggressive the budget is, and we state VarRate's operating range accordingly rather
than claiming a universal aggressive-budget win. For reference, memory-matched
quantization (KIVI-2, fixed at $\approx19\%$ budget regardless of the nominal KR
target) still wins summarization even in this aggressive regime (Llama $100/35.3/28.9$,
Qwen $94.4/33.2/30.5$ at passage/gov/musique), consistent with
Appendix~\ref{app:kivi}'s finding that the quantization axis's advantage on distributed
information does not depend on how tight the rank budget is. The mechanism behind
VarRate's collapse at KR8 on Qwen is structural rather than a tuning failure: once the
per-token rank budget shrinks far enough, a shared low-rank basis eventually cannot
represent \emph{any} coded token to a useful fidelity, and every token degrades
together. Exact selection's failure mode is different in kind, not just in degree---it
always keeps some tokens at perfect fidelity and simply drops more of them as the
budget tightens, which is why it remains viable at budgets where a shared low-rank
basis is not. Whether a deployment should prefer VarRate or SnapKV at a given budget is
therefore a question of which failure mode is more tolerable for the workload, not a
question of which method is unconditionally better.

\begin{table*}[t]
\centering\small
\begin{adjustbox}{max width=\textwidth}
\begin{tabular}{llccc}
\toprule
model & method & KR8 (passage/gov/musique) & KR10 (passage/gov/musique) & KR12 (passage/gov/musique) \\
\midrule
Llama & SnapKV & 100 / 25.0 / 27.0 & 100 / 26.0 / 28.7 & 99.5 / 26.8 / 29.7 \\
Llama & \textbf{VarRate} & 94 / 24.6 / 20.8 & 99.5 / 25.5 / 28.7 & 99.5 / 26.7 / 32.0 \\
Llama & flat & 90 / 24.0 / 16.5 & 87 / 25.1 / 18.8 & 86 / 26.7 / 23.7 \\
Qwen & SnapKV & \textbf{99.5} / 25.6 / 24.6 & 99.5 / 26.6 / 24.7 & 99.5 / 27.3 / 25.2 \\
Qwen & \textbf{VarRate} & 29.6 / 20.6 / 4.2 & 98.8 / 27.4 / 19.0 & 99.5 / 28.8 / 25.0 \\
Qwen & flat & 27.8 / 20.5 / 4.2 & 37.3 / 22.9 / 4.2 & 46.0 / 24.8 / 4.3 \\
\bottomrule
\end{tabular}
\end{adjustbox}
\caption{Aggressive-budget sweep (passage/gov/musique, query-aware, $n{=}200$). On
Qwen at KR8, VarRate collapses to flat's level---a viability floor selection does not
have. Parity returns by KR10 (Llama) / KR12 (Qwen).}
\label{tab:aggbudget}
\end{table*}

Crossing this with the stride/floor knobs at KR10 (Table~\ref{tab:aggfloor}) extends
App.~\ref{app:floor}'s finding: $r_{\min}{=}0$ still ties $r_{\min}{=}16$ at an
aggressive budget, and wider stride still helps, exactly as at KR20.

\begin{table}[tb]
\centering\small
\begin{adjustbox}{max width=\columnwidth}
\begin{tabular}{lccc}
\toprule
config (KR10, Llama) & passage & gov & musique \\
\midrule
flat & 87.0 & 25.1 & 18.8 \\
VarRate (default: s16, $r_{\min}{=}16$) & 99.5 & 25.5 & 28.7 \\
VarRate, $r_{\min}{=}0$ & 99.5 & 25.5 & 29.2 \\
VarRate, stride 64 & 99.5 & 26.2 & 30.4 \\
VarRate, $r_{\min}{=}0$ + stride 64 & \textbf{99.5} & \textbf{26.3} & \textbf{30.8} \\
\bottomrule
\end{tabular}
\end{adjustbox}
\caption{Floor and stride crossed at an aggressive KR10 budget. Both App.~\ref{app:floor}
and App.~\ref{app:stride}'s findings extend unchanged: the floor stays slack, wider
stride still helps.}
\label{tab:aggfloor}
\end{table}

\begin{figure}[tb]
\centering
\includegraphics[width=\columnwidth]{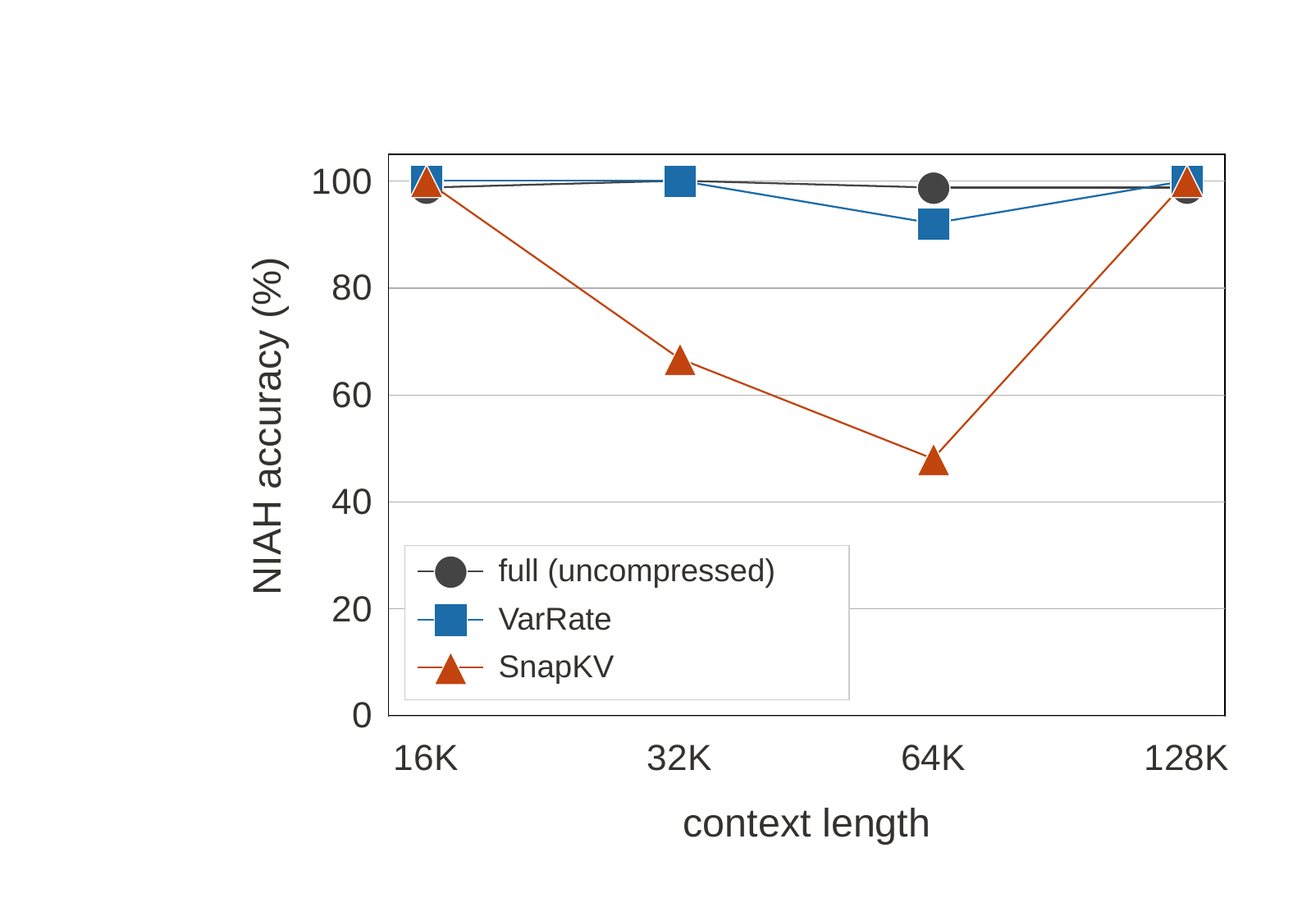}
\caption{Query-agnostic needle-in-a-haystack, Llama-3.1-8B at a matched $20\%$ budget,
fixed needle set at every length ($n{=}75$/cell). VarRate tracks the uncompressed
ceiling; SnapKV's accuracy is non-monotonic and drops to $48\%$ at $64$K. The $128$K
recovery is genuine in the raw generations but its mechanism is not established (below).}
\label{fig:niah}
\end{figure}

\subsection{Long-Context Retrieval Beyond the 16K Ceiling}
\label{app:niah}
Every accuracy result in the main text is measured at LongBench's native length
($\lesssim 16$K). We extend the query-agnostic setting to $128$K on Llama using
kvpress's needle-in-a-haystack harness, at the same $20\%$ budget (Figure~\ref{fig:niah}).
VarRate tracks the uncompressed ceiling at every length, degrading to $92\%$ at its
single worst point ($64$K) against SnapKV's $48\%$ there (Table~\ref{tab:niah}); Figure~\ref{fig:walkthrough}
traces a single needle through the codec to show why.
SnapKV's accuracy is also non-monotonic---it recovers to $100\%$ at $128$K---and a
controlled re-run with the identical needle set at every length shows this is not a
sampling artifact: the pattern survives. We report the $64$K dip and $128$K recovery as
real and reproducible (verified against the raw generations) without a confirmed
mechanism; the load-bearing claim is that VarRate degrades far less than selection
under stress at matched memory, which the control does not disturb. Realizing the
memory saving at $128$K at run time still requires the fused kernel or a multi-GPU
shard, as in \S\ref{sec:cost}.

Two further checks separate the genuine effect from possible instrumentation
artifacts. First, $128$K required sharding the model across two GPUs, because the
reconstruct-full-K/V harness (which every accuracy number in this paper uses, by
convention with prior low-rank work) exhausts a single 40--48GB GPU's memory at that
length; a dedicated shard-check confirmed that the sharded configuration reproduces
the single-GPU $32$K/$64$K numbers exactly, so the $128$K row is a deployment
consideration---realizing the accuracy requires either the fused kernel or a second
GPU at that length---rather than a signal that sharding itself changes the accuracy
result. Second, the uncompressed model's own rare misses ($\sim\!1\%$) are themselves
a decoding quirk rather than a retrieval failure: greedy decoding occasionally
truncates the six-digit needle mid-generation for the exact same needle value at
different lengths and depths (e.g.\ producing ``415.'' in place of ``915965''),
independent of any compression, which sets the honest noise floor against which
VarRate's and SnapKV's compressed-model drops should be read. Examining the raw
per-depth predictions further shows that VarRate's $64$K dip is narrow---$60\%$ at a
single depth, $100\%$ at every other depth tested---while SnapKV's dip at the same
length is broad, spanning three of five depths at $0$--$27\%$ each; both compressed
methods return to $100\%$ at $128$K, and inspection of the raw generations confirms
these are genuine correct completions rather than a scoring artifact.

\begin{table}[tb]
\centering\small
\begin{adjustbox}{max width=\columnwidth}
\begin{tabular}{lcccc}
\toprule
method @KR20 ($n{=}75$/cell) & 16K & 32K & 64K & 128K \\
\midrule
full (uncompressed) & 98.7 & 100 & 98.7 & 98.7 \\
\textbf{VarRate} & \textbf{100} & \textbf{100} & \textbf{92.0} & \textbf{100} \\
SnapKV & 100 & 66.7 & 48.0 & 100 \\
\bottomrule
\end{tabular}
\end{adjustbox}
\caption{Query-agnostic NIAH, fixed needle set at every length. VarRate's worst point
(64K, 92\%) far exceeds SnapKV's (48\%); the 128K recovery is genuine but unexplained.}
\label{tab:niah}
\end{table}


\end{document}